\makeatletter\def\graphicscache@inhibit{true}\makeatother
\makeatletter\def\includesupp{true}\makeatother
\documentclass[conference]{IEEEtran}
\usepackage{times}

\usepackage[numbers, compress]{natbib}
\usepackage{multicol}
\usepackage[bookmarks=true,hidelinks]{hyperref}

\usepackage{epsfig}
\usepackage{graphicx}
\usepackage{amsmath}
\usepackage{amssymb}

\usepackage{nccmath}
\usepackage[font=small]{caption}
\DeclareMathOperator*{\argmax}{arg\,max}

\usepackage{tikz}
\usetikzlibrary{positioning,3d,calc,intersections,spy,matrix,shapes}
\usetikzlibrary{quotes,arrows.meta}
\usetikzlibrary{positioning}

\tikzset{Ball/.pic={\tikzset{/sphere/.cd,#1}	 	

\pgfmathsetmacro{\r}{\radius*\scale}

\shade[ball color=\fill,opacity=\opacity] (0,0,0) circle (\r);
\draw (0,0,0) circle [radius=\r] node[scale=4*\r] {\logo};

\coordinate (\name-anchor) at ( 0 , 0  , 0) ;
\coordinate (\name-east)   at ( \r, 0  , 0) ;
\coordinate (\name-west)   at (-\r, 0  , 0) ;
\coordinate (\name-north)  at ( 0 , \r , 0) ;
\coordinate (\name-south)  at ( 0 , -\r, 0) ;

\path (\name-south) + (0,-20pt) coordinate (caption-node) 
edge ["\textcolor{black}{\bf \xcaption}"'] (caption-node); %

},
/sphere/.search also={/tikz},
/sphere/.cd,
radius/.store       in=\radius,
scale/.store        in=\scale,
caption/.store      in=\xcaption,
name/.store         in=\name,
fill/.store         in=\fill,
logo/.store         in=\logo,
opacity/.store      in=\opacity,
logo=$\Sigma$,
fill=green,
opacity=0.10,
scale=0.2,
radius=0.5,
caption=,
name=,
}

\tikzset{Box/.pic={\tikzset{/boxblock/.cd,#1}
        \tikzstyle{box}=[every edge/.append style={pic actions, densely dashed, opacity=.7},fill opacity=\opacity, pic actions,fill=\fill]
        
        \pgfmathsetmacro{\y}{\cubey*\scale}
        \pgfmathsetmacro{\z}{\cubez*\scale}

        \foreach[count=\i,%
                 evaluate=\i as \xlabel using {array({\boxlabels},\i-1)},%
                 evaluate=\unscaledx as \k using {\unscaledx*\scale+\prev}, remember=\k as \prev (initially 0)] 
                 \unscaledx in \cubex
        {
            \pgfmathsetmacro{\x}{\unscaledx*\scale}
            \coordinate (a) at (\k-\x , \y/2 , \z/2); 
            \coordinate (b) at (\k-\x ,-\y/2 , \z/2); 
            \coordinate (c) at (\k    ,-\y/2 , \z/2); 
            \coordinate (d) at (\k    , \y/2 , \z/2); 
            \coordinate (e) at (\k    , \y/2 ,-\z/2); 
            \coordinate (f) at (\k    ,-\y/2 ,-\z/2); 
            \coordinate (g) at (\k-\x ,-\y/2 ,-\z/2); 
            \coordinate (h) at (\k-\x , \y/2 ,-\z/2); 
        
            \draw [box] 
                (d) -- (a) -- (b) -- (c) -- cycle     
                (d) -- (a) -- (h) -- (e) -- cycle
                (f) edge (g)
                (b) edge (g)
                (h) edge (g)    
            ;
            \path (b) edge ["\xlabel"',midway] (c);
            
            \xdef\LastEastx{\k} %
        }%
        \draw [box] (d) -- (e) -- (f) -- (c) -- cycle; %
        
        \coordinate (a1) at (0 , \y/2 , \z/2);
        \coordinate (b1) at (0 ,-\y/2 , \z/2);
        \tikzstyle{depthlabel}=[pos=0,text width=14*\z,text centered,sloped]       
        
        \path (c) edge ["\tiny\zlabel"',depthlabel](f); %
        \path (b1) edge ["\ylabel",midway] (a1);  %

        \tikzstyle{captionlabel}=[text width=15*\LastEastx/\scale,text centered]       
        \path (\LastEastx/2,-\y/2,+\z/2) + (0,-12pt) coordinate (\name-cap)
        edge ["\textcolor{black}{ \bf \xcaption}"',captionlabel](\name-cap) ; %

        \coordinate (\name-west)   at (0,0,0) ;
        \coordinate (\name-east)   at (\LastEastx, 0,0) ;
        \coordinate (\name-north)  at (\LastEastx/2,\y/2,0);
        \coordinate (\name-south)  at (\LastEastx/2,-\y/2,0);       
        \coordinate (\name-anchor) at (\LastEastx/2, 0,0) ;
        
        \coordinate (\name-near) at (\LastEastx/2,0,\z/2);
        \coordinate (\name-far)  at (\LastEastx/2,0,-\z/2);       
        
        \coordinate (\name-nearwest) at (0,0,\z/2);
        \coordinate (\name-neareast) at (\LastEastx,0,\z/2);
        \coordinate (\name-farwest)  at (0,0,-\z/2);
        \coordinate (\name-fareast)  at (\LastEastx,0,-\z/2);
        
        \coordinate (\name-northeast) at (\name-north-|\name-east);
        \coordinate (\name-northwest) at (\name-north-|\name-west);
        \coordinate (\name-southeast) at (\name-south-|\name-east);
        \coordinate (\name-southwest) at (\name-south-|\name-west);
        
        \coordinate (\name-nearnortheast)  at (\LastEastx, \y/2, \z/2);
        \coordinate (\name-farnortheast)   at (\LastEastx, \y/2,-\z/2);
        \coordinate (\name-nearsoutheast)  at (\LastEastx,-\y/2, \z/2);
        \coordinate (\name-farsoutheast)   at (\LastEastx,-\y/2,-\z/2);
        
        \coordinate (\name-nearnorthwest)  at (0, \y/2, \z/2);
        \coordinate (\name-farnorthwest)   at (0, \y/2,-\z/2);
        \coordinate (\name-nearsouthwest)  at (0,-\y/2, \z/2);
        \coordinate (\name-farsouthwest)   at (0,-\y/2,-\z/2);
        
    },
    /boxblock/.search also={/tikz},
    /boxblock/.cd,
    width/.store        in=\cubex,
    height/.store       in=\cubey,
    depth/.store        in=\cubez,
    scale/.store        in=\scale,
    xlabel/.store       in=\boxlabels,
    ylabel/.store       in=\ylabel,
    zlabel/.store       in=\zlabel,
    caption/.store      in=\xcaption,
    name/.store         in=\name,
    fill/.store         in=\fill,
    opacity/.store      in=\opacity,
    fill={rgb:red,5;green,5;blue,5;white,15},
    opacity=0.4,
    width=2,
    height=13,
    depth=15,
    scale=.2,
    xlabel={{"","","","","","","","","",""}},
    ylabel=,
    zlabel=,
    caption=,
    name=,
}

\tikzset{RightBandedBox/.pic={\tikzset{/block/.cd,#1}
                
        \tikzstyle{box}=[every edge/.append style={pic actions, densely dashed, opacity=.7},fill opacity=\opacity, pic actions,fill=\fill]
        
        \tikzstyle{band}=[every edge/.append style={pic actions, densely dashed, opacity=.7},fill opacity=\bandopacity, pic actions,fill=\bandfill,draw=\bandfill]
        
        \pgfmathsetmacro{\y}{\cubey*\scale}
        \pgfmathsetmacro{\z}{\cubez*\scale}

        \foreach[count=\i,%
                 evaluate=\i as \xlabel using {array({\boxlabels},\i-1)},%
                 evaluate=\unscaledx as \k using {\unscaledx*\scale+\prev}, remember=\k as \prev (initially 0)] 
                 \unscaledx in \cubex
        {
            \pgfmathsetmacro{\x}{\unscaledx*\scale}
            \coordinate (a)     at (\k-\x   , \y/2 , \z/2); 
            \coordinate (art)   at (\k-\x/3 , \y/2 , \z/2); %
            \coordinate (b)     at (\k-\x   ,-\y/2 , \z/2); 
            \coordinate (brt)   at (\k-\x/3 ,-\y/2 , \z/2); %
            \coordinate (c)     at (\k      ,-\y/2 , \z/2); 
            \coordinate (d)     at (\k      , \y/2 , \z/2); 
            \coordinate (e)     at (\k      , \y/2 ,-\z/2); 
            \coordinate (f)     at (\k      ,-\y/2 ,-\z/2); 
            \coordinate (g)     at (\k-\x   ,-\y/2 ,-\z/2); 
            \coordinate (h)     at (\k-\x   , \y/2 ,-\z/2); 
            \coordinate (hrt)   at (\k-\x/3 , \y/2 ,-\z/2); %

            \draw [box] 
                (d) -- (a) -- (b) -- (c) -- cycle     
                (d) -- (a) -- (h) -- (e) -- cycle;
            \draw [box]
                (f) edge (g)
                (b) edge (g)
                (h) edge (g);
            \draw [band] 
                (d) -- (art) -- (brt) -- (c) -- cycle     
                (d) -- (art) -- (hrt) -- (e) -- cycle;
            \draw [box,fill opacity=0] 
                (d) -- (a) -- (b) -- (c) -- cycle     
                (d) -- (a) -- (h) -- (e) -- cycle;            
            	
            \path (b) edge ["\xlabel"',midway] (c);
            
            \xdef\LastEastx{\k} %
        }%
        \draw [box] (d) -- (e) -- (f) -- (c) -- cycle; %
        \draw [band] (d) -- (e) -- (f) -- (c) -- cycle; %
        \draw [pic actions] (d) -- (e) -- (f) -- (c) -- cycle; %
        
        \coordinate (a1) at (0 , \y/2 , \z/2);
        \coordinate (b1) at (0 ,-\y/2 , \z/2);
        \tikzstyle{depthlabel}=[pos=0,text width=14*\z,text centered,sloped]       
        
        \path (c) edge ["\tiny\zlabels"',depthlabel](f); %
        \path (b1) edge ["\ylabel",midway] (a1);  %
        
        \tikzstyle{captionlabel}=[text width=15*\LastEastx/\scale,text centered] 
        \path (\LastEastx/2,-\y/2,+\z/2) + (0,-12pt) coordinate (\name-cap)
        edge ["\textcolor{black}{ \bf \xcaption}"',captionlabel] (\name-cap); %

        \coordinate (\name-west)   at (0,0,0) ;
        \coordinate (\name-east)   at (\LastEastx, 0,0) ;
        \coordinate (\name-north)  at (\LastEastx/2,\y/2,0);
        \coordinate (\name-south)  at (\LastEastx/2,-\y/2,0);       
        \coordinate (\name-anchor) at (\LastEastx/2, 0,0) ;
        
        \coordinate (\name-near) at (\LastEastx/2,0,\z/2);
        \coordinate (\name-far)  at (\LastEastx/2,0,-\z/2);       
        
        \coordinate (\name-nearwest) at (0,0,\z/2);
        \coordinate (\name-neareast) at (\LastEastx,0,\z/2);
        \coordinate (\name-farwest)  at (0,0,-\z/2);
        \coordinate (\name-fareast)  at (\LastEastx,0,-\z/2);
        
        \coordinate (\name-northeast) at (\name-north-|\name-east);
        \coordinate (\name-northwest) at (\name-north-|\name-west);
        \coordinate (\name-southeast) at (\name-south-|\name-east);
        \coordinate (\name-southwest) at (\name-south-|\name-west);
        
        \coordinate (\name-nearnortheast)  at (\LastEastx, \y/2, \z/2);
        \coordinate (\name-farnortheast)   at (\LastEastx, \y/2,-\z/2);
        \coordinate (\name-nearsoutheast)  at (\LastEastx,-\y/2, \z/2);
        \coordinate (\name-farsoutheast)   at (\LastEastx,-\y/2,-\z/2);
        
        \coordinate (\name-nearnorthwest)  at (0, \y/2, \z/2);
        \coordinate (\name-farnorthwest)   at (0, \y/2,-\z/2);
        \coordinate (\name-nearsouthwest)  at (0,-\y/2, \z/2);
        \coordinate (\name-farsouthwest)   at (0,-\y/2,-\z/2);
    },
    /block/.search also={/tikz},
    /block/.cd,
    width/.store        in=\cubex,
    height/.store       in=\cubey,
    depth/.store        in=\cubez,
    scale/.store        in=\scale,
    xlabel/.store       in=\boxlabels,
    ylabel/.store       in=\ylabel,
    zlabel/.store       in=\zlabels,
    caption/.store      in=\xcaption,
    name/.store         in=\name,
    fill/.store         in=\fill,
    bandfill/.store     in=\bandfill,
    opacity/.store      in=\opacity,
    bandopacity/.store  in=\bandopacity,
    fill={rgb:red,5;green,5;blue,5;white,15},
    bandfill={rgb:red,5;green,5;blue,5;white,5},
    opacity=0.4,
    bandopacity=0.6,
    width=2,
    height=13,
    depth=15,
    scale=.2,
    xlabel={{"","","","","","","","","",""}},
    ylabel=,
    zlabel=,
    caption=,
    name=,
}

\usepackage{pgfplots}
\pgfplotsset{compat=1.15}
\usepgfplotslibrary{colorbrewer}
\usepgfplotslibrary{units}

\makeatletter

\tikzdeclarecoordinatesystem{rel}{%
	\tikzset{cs/.cd,x=0pt,y=0pt,#1}%
	\pgfpointlineattime{(\tikz@cs@x / 100)}%
	{\pgfpointanchor{\tikz@pp@name{\tikz@cs@node}}{south west}}%
	{\pgfpointanchor{\tikz@pp@name{\tikz@cs@node}}{south east}}%
	\edef\tikz@cs@x{\the\pgf@x}%
	\pgfpointlineattime{(\tikz@cs@y / 100)}%
	{\pgfpointanchor{\tikz@pp@name{\tikz@cs@node}}{south west}}%
	{\pgfpointanchor{\tikz@pp@name{\tikz@cs@node}}{north west}}%
	\pgfpoint{\tikz@cs@x}{\pgf@y}%
}

\makeatother

\usepackage{adjustbox}

\usepackage{booktabs}
\usepackage{threeparttable}
\usepackage{multirow}

\usepackage[capitalize]{cleveref}
\crefname{section}{Sec.}{Sections}

\usepackage{amsmath}
\usepackage{tensor}
\usepackage{float}
\usepackage{dblfloatfix}
\usepackage{lipsum}

\usepackage{blindtext,letltxmacro,xcolor,xparse}
\LetLtxMacro{\blindtextblindtext}{\blindtext}
\LetLtxMacro{\blindtextBlindtext}{\Blindtext}

\RenewDocumentCommand{\blindtext}{O{\value{blindtext}}}{%
	\begingroup\color{gray}\blindtextblindtext[#1]\endgroup
}

\IfFileExists{graphicscache.sty}{
	\ifdefined\includesupp
	\usepackage[dpi=1200]{graphicscache}
	\else
	\usepackage[dpi=1200]{graphicscache}
	\fi
}

\pgfplotstableset{search path={.,supplementary}}

\usepackage[nomargin,inline,draft]{fixme}
\fxsetup{theme=color}

\newlength{\imgheight}

\begin{document}

	\title{FaDIV-Syn: Fast Depth-Independent View Synthesis \\ using Soft Masks and Implicit Blending}
\author{Andre Rochow\\
	University of Bonn\\
	{\tt\small rochow@ais.uni-bonn.de}
	\and
	Max Schwarz\\
	University of Bonn\\
	{\tt\small schwarz@ais.uni-bonn.de}
	\and
	Michael Weinmann\\
	Delft University of Technology
	\and
	Sven Behnke\\
	University of Bonn
}

	\maketitle

	\begin{abstract}
        Novel view synthesis is required in many robotic applications, such as VR teleoperation and scene reconstruction.
        Existing methods are often too slow for these contexts, cannot handle dynamic scenes, and are limited
        by their explicit depth estimation stage, where incorrect depth predictions can lead to large projection errors.
        Our proposed method runs in real time on live streaming data and avoids explicit depth estimation by
        efficiently warping input images into the target frame for a range of assumed depth planes.
		The resulting plane sweep volume (PSV) is directly fed into our network,
		which first estimates soft PSV masks in a self-supervised manner, and then directly produces
		the novel output view.
		This improves efficiency and performance on transparent, reflective, thin, and feature-less scene parts.
		\mbox{FaDIV-Syn} can perform both interpolation and extrapolation tasks at 540p in real-time and outperforms state-of-the-art extrapolation methods on the large-scale RealEstate10k dataset.
		We thoroughly evaluate ablations, such as removing the Soft-Masking network, training from fewer examples as well as generalization to higher resolutions and stronger depth discretization. Our implementation is available\footnote{
		\url{https://github.com/AIS-Bonn/fadiv-syn}}.
	\end{abstract}

	\section{Introduction}
	
	Novel view synthesis (NVS) aims to estimate images from novel viewpoints of a scene from images
	captured from one or more reference views.
	
\begin{figure}
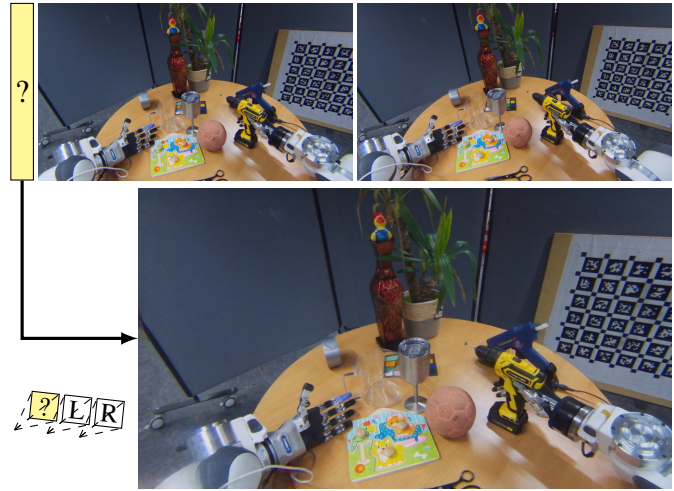

	 \centering \newlength{\teaserh}\setlength{\teaserh}{2.35cm}
	 \begin{tikzpicture}[a/.style={inner sep=0cm}]
	  \node[fill=yellow!50,draw=black,a,minimum height=\teaserh,font=\large,minimum width=0.3cm]  (placeholder) {?};
	  \node[a,right=0.05cm of placeholder]  (left) {\includegraphics[height=\teaserh]{images/robot/left.jpeg}};
	  \node[a,right=0.05cm of left] (right) {\includegraphics[height=\teaserh]{images/robot/right.jpeg}};
	  \node[a,anchor=north east] at ($(right.south east)+(0,-0.1)$) (interpolated) {\includegraphics[width=0.8\linewidth]{images/robot/extrapolated.png}};

	  \draw[very thick, -latex] (placeholder) |- (interpolated);

	  \begin{scope}[shift={(-0.1, -4.5)}, x={(0.9, -0.1)}, y={(-0.1, -0.9)}, z={(1,-1)}, scale=0.4,l/.style={xslant=0.1,yslant=-0.1},d/.style={dashed}]

	   \begin{scope}[shift={(0.0, 0.0, 0.0)}]
		\draw[fill=yellow!50] (-0.5,-0.5,1) -- (0.5, -0.5, 1) -- (0.5, 0.5, 1) -- (-0.5, 0.5, 1) -- cycle;
		\draw[d] (0,0,0) -- (-0.5, -0.5, 1);
		\draw[d] (0,0,0) -- ( 0.5, -0.5, 1);
		\draw[d] (0,0,0) -- ( 0.5,  0.5, 1);
		\draw[d] (0,0,0) -- (-0.5,  0.5, 1);
		\node[l] at (0,0,1) {?};
	   \end{scope}
	   \begin{scope}[shift={(1.2, 0.0, 0.0)}]
		\draw (-0.5,-0.5,1) -- (0.5, -0.5, 1) -- (0.5, 0.5, 1) -- (-0.5, 0.5, 1) -- (-0.5, -0.5, 1);
		\draw[d] (0,0,0) -- (-0.5, -0.5, 1);
		\draw[d] (0,0,0) -- ( 0.5, -0.5, 1);
		\draw[d] (0,0,0) -- ( 0.5,  0.5, 1);
		\draw[d] (0,0,0) -- (-0.5,  0.5, 1);
		\node[l] at (0,0,1) {L};
	   \end{scope}
	   \begin{scope}[shift={(2.4, 0.0, 0.0)}]
		\draw (-0.5,-0.5,1) -- (0.5, -0.5, 1) -- (0.5, 0.5, 1) -- (-0.5, 0.5, 1) -- (-0.5, -0.5, 1);
		\draw[d] (0,0,0) -- (-0.5, -0.5, 1);
		\draw[d] (0,0,0) -- ( 0.5, -0.5, 1);
		\draw[d] (0,0,0) -- ( 0.5,  0.5, 1);
		\draw[d] (0,0,0) -- (-0.5,  0.5, 1);
		\node[l] at (0,0,1) {R};
	   \end{scope}
	  \end{scope}
	 \end{tikzpicture}
	 \caption{FaDIV-Syn extrapolates from two cameras (L,R) mounted in a stereo setup with human baseline (6.5\,cm) to a pose outside of the baseline (denoted ?). The network was trained
	 on RealEstate10k~\citep{stereo_mag} and generalizes to robotic teleoperation scenarios.}
	 \label{fig:teaser}
	\end{figure}

	This is of great relevance for numerous applications in virtual reality, 3D movies, computer games, and other areas where images have to be generated efficiently under arbitrarily chosen viewpoints.
	VR teleoperation of robots is a specific application (see \cref{fig:teaser}). Here NVS can be employed to reduce latencies (e.g., instantaneously reacting to the viewpoint changes by the operator),
	to compensate camera workspace limits, or to generate third-person views.
	However, this application places special constraints on the NVS algorithm:
	Movement latency needs to be minimal to avoid VR sickness, i.e. novel views have to be generated in real time.
	On the other hand, scene latency, i.e. the time from image capture to display also needs to be minimal
	to allow efficient telemanipulation.
	This effectively rules out approaches which have a costly preprocessing step that generates a suitable representation
	for later interpolation.
	Note that NVS for VR teleoperation is of high importance for recently emerging Avatar systems \citep{whitney2020comparing,schwarz2021nimbro}.

	However, synthesizing novel views of a sparsely captured scene is challenging, as scene geometry and surface properties are unknown a priori and have to be inferred from the input views. Additionally, viewpoint changes
	induce both occlusions, where foreground objects occlude previously visible
	backgrounds, and disocclusions, which uncover previously invisible backgrounds.
	In the latter case, the disoccluded content has to be guessed from its context.

	Our work is focused on the problem setting of real-time view interpolation and extrapolation
	from two RGB images, which show the scene from roughly the same direction---as, for example, when captured
	from a stereo camera mounted on a robotic system. We note, though, that we also show applicability to more input views.

	Many stereo \citep{DPSNet, psv_cross_ratio, DeepMVS, thin_volumes} and view synthesis~\citep{stereo_mag, multi_mpi, kalantri} methods use Plane Sweep Volumes~(PSVs)~\citep{collins1996space},
	which warp the input views on a range of parallel planes
	and thus pre-transform the input data under the assumption of a range of discrete depths.
	Usually, a disparity or depth map is estimated from the PSV, which is then used
	to project the input views into the target frame~\citep{kalantri} or to generate representations such as multi-plane images~\citep{stereo_mag, multi_mpi}.
	This approach exhibits a fundamental bottleneck, however: Imprecise or wrong depth estimates, which
	occur especially on uniform, transparent, thin, or reflective surfaces, will
	result in loss of information and lead to failures later on in the synthesis pipeline.
	
	Our proposed method is related to Image-Based Rendering (IBR) approaches, but forgoes the geometry estimation step by directly operating on the PSV in the target view to produce the output RGB image, without computing explicit depth.
	To this end, we learn an RGB generator network which processes the PSV to directly synthesize the novel view and equip it with operations
	such as group convolutions and gated convolutions~\citep{gated_convs, generative_inpainting},
	which are suitable for detecting layer-wise correspondences, masking of irrelevant areas within the layers, and blending.
	Unlike most IBR approaches, FaDIV-Syn has no explicit blending or inpainting stage, which avoids early and hard decisions, allowing the
	distribution of the blending and inpainting operations throughout the learned network.
	Only a single forward pass is required for view interpolation and extrapolation.
	Further, we demonstrate how gated convolutions can be used to learn self-supervised soft masks, which are predicted by a lightweight CNN to give a soft correspondence measure, which improves accuracy
	and provides insight on the network's method of operation.

	In summary, our contributions include 1) a real-time view synthesis method operating on plane sweep volumes without requiring explicit geometry, 2) self-supervised learning of soft-masks by introducing a gating module, and 3) a detailed evaluation on the large-scale RealEstate10k dataset~\citep{stereo_mag},
	where we demonstrate our approach to outperform existing methods in terms of accuracy and runtime.

	\section{Related Work}
	Novel view synthesis has gained much attention in recent years and a large variety of approaches has been presented. For a broader review, we refer to surveys by \citet{nguyen2020rgbd} and \citet{tewari2020state}.

	\paragraph{Image Based Rendering (IBR)}
	In contrast to classical rendering of 3D scenes using textured geometry, IBR methods render novel views by combining input images in the target pose~\citep{soft3d,DeepBlending, ibr_scaleable, selective_ibr, ibrnet,kalantri,nguyen2020rgbd,Deep3D,thies2020imageguided, riegler2020free, riegler2020stable, deep_stereo}.
	To be able to project the input images correctly, IBR methods still require geometry, often in
	the form of depth maps, which are either available or estimated.
	Recent approaches use blending to combine the images~\cite{soft3d,DeepBlending, ibr_scaleable, riegler2020free}.
	\Citet{DeepBlending} learn the blending operation end-to-end.
	\Citet{riegler2020free} use a recurrent blending decoder in order to deal with a varying number of input images.
	In recent work~\citep{riegler2020stable}, heuristic input image selection is replaced with a fully-differentiable synthesis block.
	\Citet{soft3d} introduce soft visibility volumes, which encode occlusion probabilities and thus avoid early
	decisions, but require a larger number of input views.
	\Citet{kalantri} synthesize novel views in light field datasets. They use the corner cameras to predict depth in the target view, which is then used to warp the input views.
	\Citet{nguyen2020rgbd} introduce \mbox{RGBD-Net}, which first estimates depth using a multi-scale PSV,
	warps the input images into the target frame, performs explicit blending, and refines the warped image
	using a depth-aware network.
	Similar to our approach, \citet{deep_stereo} also directly build a PSV in the target view.
	Color is fused for each PSV plane pair and a separate network estimates depth probabilities for blending.
	Our method also works directly on the input images, but does not compute or require depth explicitly.
	Blending is learned implicitly by the network, together with detection and inpainting of extrapolated/disoccluded regions.

	\paragraph{Geometry-based Approaches}
	Recent geometry-based approaches \cite{point_based_graphics,extremeview, synsin, lightfield_raydepth} use depth features to spatially project pixel information and refine these projections to a target view.
	\citet{synsin} and \citet{mono_ibr} process single input images, estimating monocular depth in an end-to-end fashion. \citet{synsin} implement a differentiable point cloud renderer that allows z-buffering and splatting. \citet{mono_ibr} predict depth in the target view using a transforming auto-encoder that explicitly transforms the latent code before entering the decoder. Similarly, \citet{tbn} learn implicit voxel representations and transform encoded representations explicitly.
	\Citet{lightfield_raydepth} predict RGB-D light fields from a single RGB image. They estimate precise scene geometry, render it to the target frame, and predict occluded rays using a second CNN.
	\citet{extremeview} predict depth probabilities along camera rays in multiple input images and unite them in the target camera pose.
	They discretize the number of possible depth values and therefore reduce the depth estimation problem to a classification problem. 
	A more recent approach~\cite{hani2020continuous} learns novel view synthesis without target view supervision
	by performing two synthesis steps, initially to an arbitrary target pose and from there to a pose where ground truth is available.
	In contrast to these methods, FaDIV-Syn does not feature an explicit geometry representation.
	We argue that explicit geometry---besides requiring a higher computational effort---forces early resolution of
	ambiguities, which can lead to loss of information.

	\begin{figure*}
		\centering
		\includegraphics[width=0.9\textwidth]{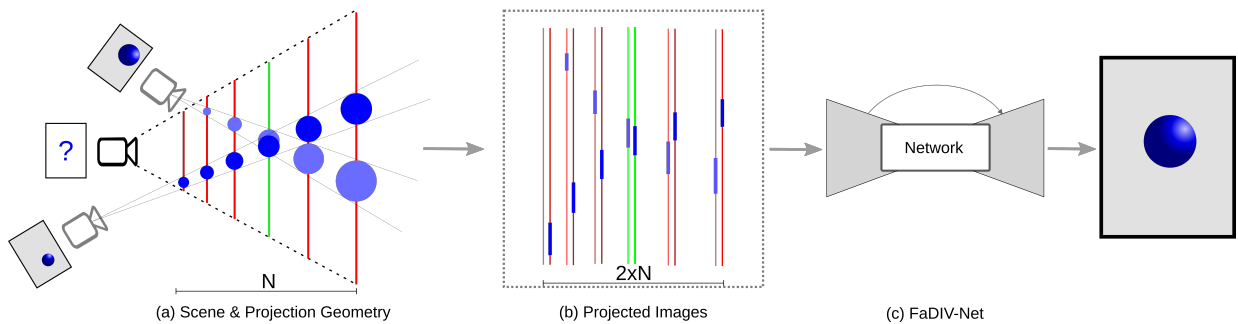}
		\caption{The FaDIV-Syn architecture.
			(a) Input images (gray) are projected into the target camera (black) for each depth plane (red/green) defined in the target frame.
			For a particular surface in the scene (blue circle), there will be a depth plane where
			the projections most closely align (green). This plane corresponds to the approximate object depth.
			(b) The resulting projected images are stacked and fed into the view synthesis network (c), which
			directly predicts the target image.}
		\label{fig:architecture} %
	\end{figure*}

	\paragraph{Multiplane and Layered Depth Images (MPIs/LDIs)}
	A multiplane image consists of multiple depth planes, which store RGB and alpha values. Once computed for a set of input images, novel views can be synthesized very efficiently by warping and blending the individual layers.
	One can attempt to predict MPIs from single input images~\cite{adaptive_mpi,single_view_mpi}.
	\Citet{single_view_mpi} train a network to estimate scale-invariant depth and require
	additional sparse point clouds to recover scale.
	Multiview approaches~\cite{multi_mpi,stereo_mag, llff} use information from additional camera poses to place surfaces at the correct MPI layer. Plane sweeping~\cite{stereo_mag} or warping~\cite{multi_mpi} at different depths creates a suitable representation for the network.
	\Citet{llff} blend the layers of multiple MPIs to generate novel views with local light fields.
	Recently, \citet{msi} extended the key idea of multiplane images to multi-sphere images in order to synthesize $360^\circ$ images in real-time, however, at lower resolution.
	
	Other approaches~\cite{ldi_view,3D_photo,video_interpolation_ldi} build on Layered Depth Images (LDI)~\cite{ldi}, which store multiple RGB and depth values per pixel.
	\Citet{ldi_view} train a CNN to predict a two-layered LDI, where the network learns to predict occluded pixels.
	\Citet{3D_photo} use an LDI representation to turn a single image into a 3D photo by inpainting color and depth of the occluded areas.
	While inspired by MPI approaches, FaDIV-Syn bypasses MPI generation and instead determines a novel view from multiple warped planes directly. It is applicable for dynamic scenes in real-time.

	\paragraph{Neural Rendering}
	In recent years, view synthesis approaches based on Neural Radiance Fields (NeRF)~\citep{mildenhall2020nerf}
	have been introduced, which employ a neural network as a learnable density and radiance function over the
	scene volume. Novel views can be synthesized using classical volume rendering techniques.
	The sub-sequent improvements~\citep{zhang2020nerfpp,yu2020pixelnerf,ibrnet, Park20arxiv_nerfies} show impressive results
	on a variety of scenes
	and recent developments also achieve real-time inference~\cite{reiser2021kilonerf,Garbin-2021-FastNeRF,yu2021plenoctrees,sitzmann2021lfns}.
	However, with the exception of~\citet{ibrnet} and \citet{sitzmann2021lfns}, NeRFs typically have to be trained with hundreds of images of the target scene, making
	them unsuitable for dynamic scenes.
	While methods designed for dynamic scenes exist~\citep{Park20arxiv_nerfies,Pumarola20arxiv_D_NeRF, Gafni20arxiv_DNRF, Tretschk20arxiv_NR-NeRF}, they require offline training or processing phases as well.

	\begin{figure*}
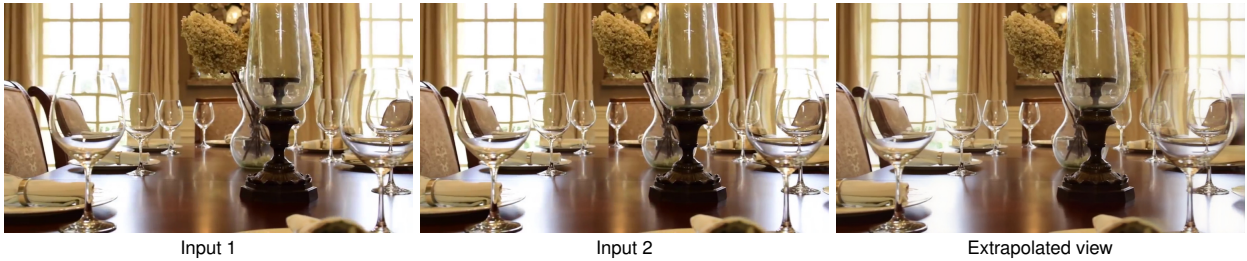

		\centering
		\sf \scriptsize \setlength{\tabcolsep}{1.5pt} \setlength{\imgheight}{3.05cm}%
		\begin{tabular}{ccc}
			\includegraphics[height=\imgheight]{images/qualitative_results/6844_ours32/helper_frame.png}
			& \includegraphics[height=\imgheight]{images/qualitative_results/6844_ours32/reference_frame.png}
			& \includegraphics[height=\imgheight]{images/qualitative_results/6844_ours32/17.png} \\
			Input 1
			& Input 2
			& Extrapolated view
		\end{tabular}
		\vspace{-0.1cm}
		\caption{View extrapolation on RealEstate10k~\citep{stereo_mag} test set. Please see our supplementary video for an animated version of this figure that shows both inter- and extrapolation along a trajectory.
		}
		\label{fig:view_extrapolation_scene}
	\end{figure*}
	
	\begin{figure*}
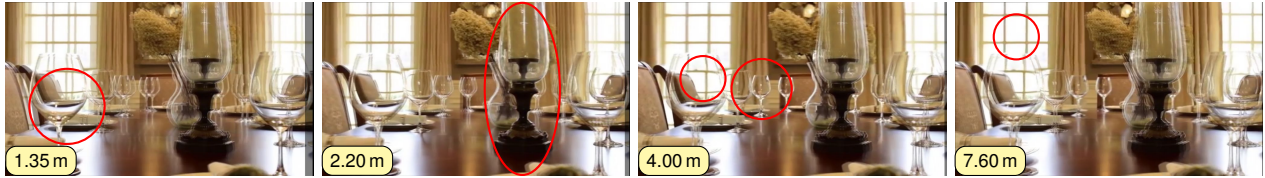
 \centering \setlength{\imgheight}{2.3cm} \small
		\tikzset{plab/.style={font=\sffamily\scriptsize, rounded corners, draw=black, fill=yellow!40}}
		\tikz{
			\node[anchor=south west, inner sep=0pt] (img) {\includegraphics[height=\imgheight]{images/psv/psv_6.jpg}};
			\draw[thick,red] (rel cs:x=20,y=40,name=img) circle (0.5cm);
			\node[anchor=south west,plab]{1.35\,m};
		}
		\tikz{
			\node[anchor=south west, inner sep=0pt] (img) {\includegraphics[height=\imgheight]{images/psv/psv_8.jpg}};
			\draw[thick,red] (rel cs:x=65,y=50,name=img) circle [x radius=0.5cm, y radius=1.145cm];
			\node[anchor=south west,plab]{2.20\,m};
		}
		\tikz{
			\node[anchor=south west, inner sep=0pt] (img) {\includegraphics[height=\imgheight]{images/psv/psv_11.jpg}};
			\draw[thick,red] (rel cs:x=40,y=50,name=img) circle [x radius=0.4cm, y radius=0.4cm];
			\draw[thick,red] (rel cs:x=21,y=56,name=img) circle [x radius=0.3cm, y radius=0.3cm];
			\node[anchor=south west,plab]{4.00\,m};
		}
		\tikz{
			\node[anchor=south west, inner sep=0pt]{\includegraphics[height=\imgheight]{images/psv/psv_15.jpg}};
			\draw[thick,red] (rel cs:x=20,y=80,name=img) circle [x radius=0.3cm, y radius=0.3cm];
			\node[anchor=south west,plab]{7.60\,m};
		}
		\caption{Plane sweep volume (PSV). Four planes of the scene in \cref{fig:view_extrapolation_scene} with $\alpha$-blended projections of the two input images. Note that RealEstate10k only provides estimated
			global scale, so the given plane distances are up to a scale.
			Exemplary areas with good correspondence between the two input views are marked in red.
			Forwarding the corresponding regions would already yield an approximate solution.}
		\label{fig:psv} \vspace{-2ex}
	\end{figure*}

	\begin{figure}[b]
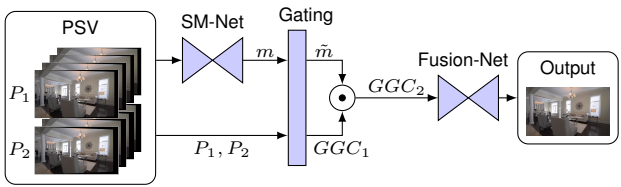

 \centering
 \begin{tikzpicture}[font=\sffamily\scriptsize,m/.style={draw=black,rounded corners},every node/.style={align=center},
    net/.style={fill=blue!20}
    ]
  \node[m,minimum height=2.3cm,minimum width=2cm] (psv) at (0,0) {};
  \node[anchor=north] at (psv.north) {PSV};
  \coordinate (psv1) at ($(psv)+(-0.05,0.05)$);
  \coordinate (psv2) at ($(psv)+(-0.05,-0.7)$);
  \node[shift={(0.3,0.3)},  draw=gray,fill=white,inner sep=0.1pt] at (psv1) {\includegraphics[width=1.1cm]{images/a035_psv/psv_ref_0.jpg}};
  \node[shift={(0.3,0.3)},  draw=gray,fill=white,inner sep=0.1pt] at (psv2) {\includegraphics[width=1.1cm]{images/a035_psv/psv_src_0.png}};
  \node[shift={(0.2,0.2)},  draw=gray,fill=white,inner sep=0.1pt] at (psv1) {\includegraphics[width=1.1cm]{images/a035_psv/psv_ref_10.jpg}};
  \node[shift={(0.2,0.2)},  draw=gray,fill=white,inner sep=0.1pt] at (psv2) {\includegraphics[width=1.1cm]{images/a035_psv/psv_src_10.png}};
  \node[shift={(0.1,0.1)},  draw=gray,fill=white,inner sep=0.1pt] at (psv1) {\includegraphics[width=1.1cm]{images/a035_psv/psv_ref_20.jpg}};
  \node[shift={(0.1,0.1)},  draw=gray,fill=white,inner sep=0.1pt] at (psv2) {\includegraphics[width=1.1cm]{images/a035_psv/psv_src_10.png}};
  \node[shift={(0.0,0.0)},  draw=gray,fill=white,inner sep=0.1pt] at (psv1) {\includegraphics[width=1.1cm]{images/a035_psv/psv_ref_30.jpg}};
  \node[shift={(0.0,0.0)},  draw=gray,fill=white,inner sep=0.1pt] at (psv2) {\includegraphics[width=1.1cm]{images/a035_psv/psv_src_30.png}};

  \node[left=0.47cm of psv1] {$P_1$};
  \node[left=0.47cm of psv2] {$P_2$};

  \coordinate (maskc) at ($(psv.east)+(0.75,0.5)$);
  \node (maskl) [net,isosceles triangle,isosceles triangle stretches,draw=black,inner sep=0pt,minimum width=0.6cm,minimum height=0.4cm,anchor=apex,outer sep=0pt] at (maskc) {};
  \node (maskr) [net,isosceles triangle,isosceles triangle stretches,draw=black,inner sep=0pt,minimum width=0.6cm,minimum height=0.4cm,anchor=apex,outer sep=0pt,rotate=180] at (maskc) {};

  \node[net,draw=black,minimum width=1.8cm,right=1.75 of psv,rotate=90,anchor=north] (ggc) {};

  \node[circle,draw=black,right=0.3 of ggc.south,inner sep=2pt] (mult) {$\cdot$};
  \draw[fill=black] (mult) circle (0.04cm);

  \coordinate (fc) at ($(mult.east)+(1.5,0)$);
  \node (fl) [net,isosceles triangle,isosceles triangle stretches,draw=black,inner sep=0pt,minimum width=0.6cm,minimum height=0.4cm,anchor=apex,outer sep=0pt] at (fc) {};
  \node (fr) [net,isosceles triangle,isosceles triangle stretches,draw=black,inner sep=0pt,minimum width=0.6cm,minimum height=0.4cm,anchor=apex,outer sep=0pt,rotate=180] at (fc) {};

  \node[m,right=0.25cm of fr.lower side] (out) {Output\\[1ex]
    \includegraphics[width=1.1cm]{images/a035_psv/pred.png}
  };

  \draw[-latex] (maskl-|psv.east) -- (maskl);
  \draw[-latex] (maskr) -- (maskr-|ggc.north) node[midway,above,inner sep=1pt] {$m$};

  \draw[-latex] ($(psv.east)+(0,-0.5)$) -- ($(ggc.north)+(0,-0.5)$) node[midway,below,inner sep=2pt] {$P_1,P_2$};

  \draw[-latex] ($(ggc.south)+(0,0.5)$) -| (mult) node[pos=0.25,above,inner sep=1pt] {$\tilde{m}$};
  \draw[-latex] ($(ggc.south)+(0,-0.5)$) -| (mult) node[pos=0.5,below,inner sep=2pt] {$GGC_1$};

  \draw[-latex] (mult) -- (fl) node[midway,above,inner sep=1pt] {$GGC_2$};
  \draw[-latex] (fr) -- (out);

  \node[above=0.3cm of maskc] {SM-Net};
  \node[above=0.3cm of fc,xshift=-2pt] {Fusion-Net};
  \node[anchor=south] at (3,0.85) {Gating};
 \end{tikzpicture}
 \caption{View synthesis network architecture. The Soft-Masking (SM) network computes layer-wise masks from the PSV, which are used
 for gating. The fusion network then produces the final output image from the gated PSV.}\label{fig:network}
\end{figure}

\newlength{\softmaskheight}\setlength{\softmaskheight}{1.01cm}
\newcommand{\mask}[1]{\includegraphics[height=\softmaskheight,clip,trim=20 10 20 10]{images/masks/60a_inverted/#1}}
\begin{figure*}
 \centering
 \begin{tikzpicture}[
   font=\sffamily\tiny,
   a/.style={inner sep=0.1pt},
   lu/.style={anchor=south,text=black,inner sep=2pt},
   ld/.style={anchor=north,text=black,inner sep=2pt}
 ]
   \node[inner sep=0pt] (rgb) {\includegraphics[height=2.22\softmaskheight,clip,trim=10 10 10 10]{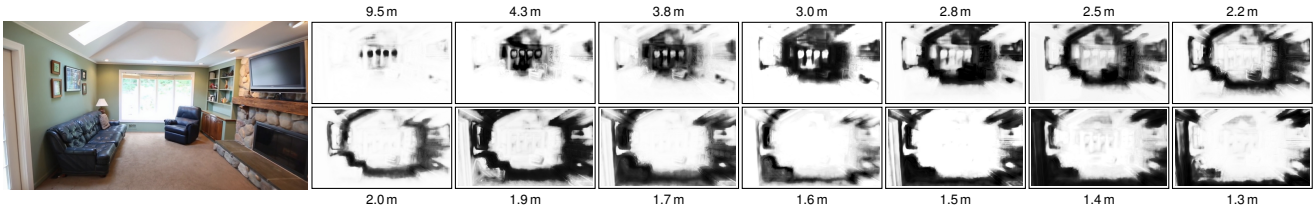}};

   \matrix (masks) [inner sep=0pt, right=1pt of rgb, matrix of nodes, every node/.style={a,draw=black, inner sep=1pt}, column sep=1pt, row sep=1pt]
   {
    |(m0)| \mask{mask_mean_3_0.png}  &
    |(m1)| \mask{mask_mean_7_4.png}  &
    |(m2)| \mask{mask_mean_8_0.png}  &
    |(m3)| \mask{mask_mean_10_0.png} &
    |(m4)| \mask{mask_mean_11_4.png} &
    |(m5)| \mask{mask_mean_12_3.png} &
    |(m6)| \mask{mask_mean_14_5.png} \\
    |(m7)| \mask{mask_mean_15_0.png} &
    |(m8)| \mask{mask_mean_16_0.png} &
    |(m9)| \mask{mask_mean_18_4.png} &
    |(m10)| \mask{mask_mean_19_5.png} &
    |(m11)| \mask{mask_mean_20_2.png} &
    |(m12)| \mask{mask_mean_22_2.png} &
    |(m13)| \mask{mask_mean_23_0.png} \\
   };

   \node[lu] at (m0.north) {9.5\,m};
   \node[lu] at (m1.north) {4.3\,m};
   \node[lu] at (m2.north) {3.8\,m};
   \node[lu] at (m3.north) {3.0\,m};
   \node[lu] at (m4.north) {2.8\,m};
   \node[lu] at (m5.north) {2.5\,m};
   \node[lu] at (m6.north) {2.2\,m};
   \node[ld] at (m7.south) {2.0\,m};
   \node[ld] at (m8.south) {1.9\,m};
   \node[ld] at (m9.south) {1.7\,m};
   \node[ld] at (m10.south) {1.6\,m};
   \node[ld] at (m11.south) {1.5\,m};
   \node[ld] at (m12.south) {1.4\,m};
   \node[ld] at (m13.south) {1.3\,m};
 \end{tikzpicture}%
 \caption{Self-supervised soft masking. Predicted image (left) and learned mask activations (right).
 The masks are normalized (dark = high activation).
 The PSV layer depth is up to scale.
 Note how the learned masks correlate with depth in the scene.}
 \label{fig:soft_masks}%
\end{figure*}
	
	\section{Proposed Method}

	The key idea of FaDIV-Syn, illustrated in \cref{fig:architecture}, is to preprocess and transform the input images into the
	target frame, without making early and hard decisions.
	Of course, this transformation requires depth information. Instead
	of explicitly estimating depth, which exhibits problems for transparent, reflective, thin, and featureless surfaces,
	we sample multiple depth values and present the resulting
	possibilities to a deep neural network for learning image synthesis. The induced multi-plane representation is well-suited for
	view synthesis.
	
	We present
	and evaluate our method for two input images (in the following denoted by $I_1$ and $I_2$) and
	one output image ($I_O$).
	This scenario commonly occurs in many contexts, such as robotics, AR/VR, and mobile devices.
	We note though, that additional input views enhance the performance
	further (see \cref{sec:eval:multiview}).
	For all input cameras, we sample $N$ depth planes and for each plane assume that the entire image lies on it.
	When projecting these planes into the target view, one could, for $N\rightarrow \infty$, determine each pixel's 3D position by searching for correspondences in the warped planes (see \cref{fig:architecture,fig:psv}).
	When performing the synthesis task with a learned network,
	we can reduce $N$ to a small number, since the network can learn to interpolate
	between planes with adjacent depth levels.

	\subsection{Plane Sweep Volumes for View Synthesis}
	
	Planar geometry is especially well-suited for camera-to-camera projection, since
	the resulting warping operation can be performed efficiently.
	We define the planes in the target image $I_O$ (see \cref{fig:architecture}),
	as it is commonly done in plane sweeping multi-view stereo approaches~\citep{collins1996space}.
	For each plane $i$ with depth $d_i$, we define $P_k^{(i)}$ as the image resulting from projecting
	$I_k$ onto the plane, and then into $I_O$. Using this representation, we can define
	a ``hard-wired'' view synthesis method $f$:
	
	\begin{alignat}{2}
	f(I_1, I_2, p) &= \begin{cases}
	P_1^{(D(p))}(p) & \text{ if } p \text{ visible in } I_1,I_2 \\
	g(I_1,I_2,p)    & \text{ otherwise },
	\end{cases} \label{eq:hardwired}\\
	D(p) &= \argmax_{j} Q(P_1^{(j)}, P_2^{(j)}, p), \label{eq:correspondence_qual}
	\end{alignat}
	where $p=(x,y)$ is a pixel in the target image $I_O$, $g$ is an inpainting method,
	$D$ is the (internal) depth estimate, $Q$ is a correspondence quality estimator,
	and $j$ denotes the plane with optimal correspondence.
	Note that only images in $P$ of the same depth need to be compared.
	\Cref{fig:psv} shows an exemplary PSV where the idea presented in \cref{eq:hardwired}
	is immediately apparent.
	Since perfect $Q$ and $g$ are, however, not known, we will learn a CNN approximating $f$.

	\paragraph{Depth Discretization}
	\label{sec:method:discretization}
	For increasingly distant objects, the spatial error of projected pixels caused by wrong depth decreases. Therefore, it is legitimate to set a maximum depth.
	Our most efficient network uses 19 depth planes, which is much less compared to related methods \cite{multi_mpi,stereo_mag} but
	achieves significantly better accuracy (see \cref{sec:eval}).
	The depths are sampled in disparity space, where we linearly interpolate within a chosen disparity range.

\subsection{View Synthesis Network}

When generating a novel view directly from a PSV, disocclusion areas are not trivial to determine.
Thus, our network must learn to distinguish between (1) areas of sufficient correspondences in the warped planes, (2) areas of no correspondence but sufficient correspondence in different planes, and (3) areas of disocclusion and occlusion.
Hence, the network must learn under the constraint of geometric consistency to fuse and correct sufficiently corresponding areas from warped planes and recognize inpainting areas to fill them with realistic content.
In contrast to \citet{thies2020imageguided}, who also learn implicit blending, we avoid early and hard
decisions in a depth estimation stage, which may lead to quality degradation later in the pipeline.

Our network is divided into several components (see \cref{fig:network}): A soft-masking network, a gating section, and the final fusion network.

\subsubsection{Soft-Masking Network}\label{sm_net}

The Soft-Masking (SM) network operates on the input PSV and generates approximate
PSV masks that guide the further synthesis process.
Ideally, the output mask $m$ should be the correspondence quality measure $\smash{Q(P_1^{(j)},P_2^{(j)})}$ (see \cref{eq:hardwired}). In practice, the entire pipeline is trained end-to-end without direct supervision on the masks.

As shown in \cref{fig:soft_masks}, the SM network learns meaningful correspondence masks that correlate closely with depth in the scene. Since there is a large amount of overlap, we conclude that the later fusion network uses the inferred masks mostly to eliminate areas of poor correspondence, but
keeps multiple possibilities to make final decisions in deeper layers.
We thus denote these representations as \textit{soft masks}.

To generate $m$, we implement an hourglass network module with four downsampling and four upsampling blocks. Each block has one convolution followed by average pooling or bilinear upsampling. The module processes the PSV ($P_1,P_2$) as $2 \times N$ input planes in grayscale and outputs $2 \times N$ feature maps. For each plane, we encourage the network to learn
a binary classification mask by applying the softmax function (separately for each plane).
For details we refer to the supplementary material.

\subsubsection{Learned Gating}

We send the entire PSV through a gating section. This way, the network can
eliminate areas of low correspondence.

Gated convolutions \cite{gated_convs} have recently shown promising results in image inpainting \cite{generative_inpainting}.
Instead of $C(W_f,I)=\sigma(W_f \circledast I)$, a gating layer calculates the non-linear output
\begin{equation}
GC(W_f, W_g, I) = \\
\sigma_1( W_f \circledast I) \odot \sigma_2(W_g \circledast I),
\end{equation}
where $\odot$ denotes the element-wise multiplication and $\circledast$ the convolution. $W_g, W_f$ are two different convolutional filters, and $\sigma_{1},\sigma_{2}$ are activation functions. The formula shows that gating directly influences the actual feature extraction and can thus adapt it to the context. Gating layers can help the network especially
with performing masking-like operations (e.g. when recognizing corresponding
depth planes), performing blending, or determining inpainting areas~\citep{generative_inpainting}.

To avoid hard decisions, we allow the gating stage to incorporate the mask $m$ in
a learned fashion by adding convolutional layers that can modify the mask and features before the actual multiplication (see \cref{fig:network}).
We note that the network may also learn to ignore the provided mask.

Since the depth planes are aligned, we can be sure that corresponding areas only appear in corresponding planes (see \cref{fig:architecture}).
Hence, we can restrict the convolutions through grouping under the assumption that other planes are initially irrelevant to the
considered plane pair.

We concatenate the estimated masks $m$ to the input plane pairs $\smash{P_1^{(j)},P_2^{(j)}}$ and activate the gating feature extraction with a sigmoid ($\sigma$), which creates a mask $\tilde{m} \in [0,1]$:

\begin{alignat}{2}
GGC_1^{(j)} &= \mathrm{ReLU}(W_1 \circledast[P_1^{(j)},P_2^{(j)},m^{(j)}]), \\
\tilde{m}^{(j)} &= \sigma(W_2 \circledast GGC_1^{(j)}), \label{eq:softmask} \\
GGC_2^{(j)} &= \mathrm{ReLU}(W_3 \circledast GGC_1^{(j)}) \odot \tilde{m}^{(j)},
\end{alignat}
where $W$ denotes the learned weights.

By allowing self-supervised learning of a suitable mask presentation and incorporation, we thus
(i) keep the key idea of our approach to not estimate depth explicitly,
(ii) keep the complexity of the mask generator bounded, and
(iii) avoid early projection errors due to the depth discretization.
	
	\subsubsection{Fusion Network} \label{fusion_net}

	Finally, the gated PSV features enter the fusion and inpainting module (Fusion-Net in \cref{fig:network}).
	It is based on a U-Net~\cite{unet} architecture and
	consists of four downsampling and four upsampling blocks with skip connections and a dilated convolution in the middle. For upsampling, we use bilinear interpolation.
	Each downsampling block consists of two convolutions with batch normalization and ReLU activation.
	The second convolution has a stride of two for downsampling.
	The final output of the fusion network, following a $\mathrm{tanh}$ activation, is the predicted RGB image $\hat{I}_O$.

	\paragraph{Variant without soft masking (NoSM)}
	To validate the effect of soft masking, we also tested generating novel views without the SM network.
	Here, the gating section receives only $P_1,P_2$ as input.
	In this setup, the network must learn to spread the correspondence estimation throughout the layers.
	We therefore use Gated Convolutions in each downsampling layer and increase the number of feature-maps in deeper layers.
	Note that both manipulations are parameter-intensive and increase the accumulated number of parameters by approximately 33\%-55\%, depending on the number of PSV planes.

	Architecture details can be found in the supplementary.

	\subsection{Training}
	
	The whole pipeline is trained end-to-end in a supervised manner from a triple $(I_1, I_2, I_O)$ with
	known camera poses and intrinsics.
	We define the loss function
	\begin{alignat}{4}
	& \mathcal{L}(\hat{I}_O, I_O) = \lambda_1 \mathcal{L}_1(\hat{I}_O, I_O) + \lambda_p \mathcal{L}_{\mathit{perceptual}}(\hat{I}_O, I_O),
	\end{alignat}
	where the perceptual loss is based on a VGG-19~\cite{vgg_19} network $\Psi$ pretrained on ImageNet. It is defined as
	\begin{alignat}{4}
	&\mathcal{L}_{\mathit{perceptual}}(\hat{I}_O, I_O) = \sum_l^L \frac{w_l}{N_{\Psi_l}}|\Psi_l(\hat{I}_O)-\Psi_l(I_O)|,
	\end{alignat}
	where $\Psi_l(\cdot)$ is the activation of the $l$-th layer, $w_l$ is a weight factor of the $l$-th layer and $N_{\Psi_l}$ are the number of elements in the $l$-th layer.
	We train the network with a batch size of 20, a learning rate of 1e-4, and the Adam optimizer with $\beta_{1,2} = (0.4,0.9)$ on two NVIDIA A6000 GPUs with 48\,GiB RAM.
	Training takes four days for images of 288p resolution. Training for a higher resolution of 576p is only possible with a batch size of six and takes up to three weeks.
	
	For 288p images, we train the networks for 300k-350k iterations, whereas for 576p images we increase the number of iterations accordingly to adjust to the smaller batch size.
	Within this range, we use early stopping based on the validation score to select the model for evaluation.
	
	\section{Evaluation} \label{sec:eval}
	
	We train and evaluate our method on the challenging RealEstate10k dataset introduced by
	\citet{stereo_mag}, which contains approx.\ 75k video clips extracted from YouTube
	videos, showing mostly indoor scenes. The large variety in the dataset allows generalizing to different scenarios (see \cref{fig:teaser}).  All videos have been automatically annotated
	with camera intrinsics and camera trajectories using ORB-SLAM2~\citep{mur2017orb} and bundle adjustment.
	Monocular SLAM cannot recover global scale, so the sequences have been scaled so that the
	near geometry lies at approx.\ 1.25\,m~\citep{stereo_mag}.
	We divide the official \textit{train} split further into 54k training and 13.5k validation sequences.
	All our tests are done using the official test split, where the extrapolation experiments use the data provided by \citet{3D_photo}.
	For all experiments, we evaluate the quality of generated images with the PSNR, SSIM~\cite{ssim}, and LPIPS~\cite{lpips} metrics.

\subsection{Extrapolation}
\label{sec:eval:extrapolation}

The first, more challenging task is extrapolation, i.e. the target view is outside of the provided input views.
We start from a pre-trained interpolation network (see \cref{sec:eval:interpolation}) and add extrapolation sequences.
Extrapolation and interpolation triplets are mixed in the ratio 80\% to 20\% during training.
These are randomly sampled from video sequences, ensuring a distance $d_1 \in [3,5]$ between
the input frames $I_0,I_1$, as well as that the target frame $I_O$ is $d_2 \in [5,7]$ frames after $I_1$.

\Citet{3D_photo} evaluated an array of related methods for extrapolation tasks. In order to test view synthesis accuracy, they generated 1500 random triplets from the test data set. We follow the same evaluation protocol.
Note that these experiments are carried out at a resolution of 1024$\times$576.

\begin {table} \centering \footnotesize  \setlength{\tabcolsep}{4pt}
\begin{tabular}{@{}l@{\hspace{4pt}}lcccrr@{\hspace{0pt}}r@{}} \toprule
	
	&            &                   &                   &                   &          & \multicolumn{2}{@{}c@{}}{Depth [m]} \\
	\cmidrule () {7-8}
	Method    &  \adjustbox{rotate=90,lap=3ex,trim=0 0 10 0}{576p}     & SSIM$\uparrow$    & PSNR$\uparrow$    & LPIPS$\downarrow$ & Params & min & max \\

	\midrule
	Stereo-Mag \cite{stereo_mag}  &\ \ \checkmark  & .8906  & 26.71            & .0826          & 17M  & 1.0 & 100 \\
	PB-MPI \cite{multi_mpi}       &\ \ \checkmark  & .8773           & 25.51            & .0902          & 6M   & 1.0 & 100 \\
	LLFF \cite{llff}              &\ \ \checkmark  & .8062           & 23.17            & .1323          & 682K & 1.0   & 100   \\
	Xview \cite{extremeview}      &\ \ \checkmark  & .8628           & 24.75            & .0822          & 58M  & -   & -   \\
	3D-Photo \cite{3D_photo}      &\ \ \checkmark  & .8887           & 27.29            & .0724          & 119M & -   & -   \\
	\midrule
	Ours-32                       &\ \ \checkmark  & \textbf{.9036}  & \textbf{29.41}   & \textbf{.0521} & 12M  & 1.0 & 100 \\
	Ours-32                       &\ \             & .9020           & 29.39            & .0556          & 12M  & 1.0 & 100 \\
	Ours-32-NoSM                  &\ \ \checkmark  & .8891           & 28.83            & .0613          & 16M  & 1.0 & 100 \\
	Ours-32-NoSM                  &\ \             & .8865           & 28.67            & .0650          & 16M  & 1.0 & 100 \\
	Ours-19                       &\ \ \checkmark  & .9007           & 29.25            & .0531          & 9M   & 1.0 & 100 \\
	Ours-19                       &\ \             & .8985           & 29.04            & .0583          & 9M   & 1.0 & 100 \\
	Ours-19-NoSM                  &\ \             & .8801           & 28.30            & .0679          & 14M  & 1.0 & 100 \\
	Ours-17-NoSM                  &\ \ \checkmark  & .8790           & 28.13            & .0674          & 14M  & 0.3 & 16  \\
	Ours-17-NoSM                  &                & .8750           & 27.96            & .0695          & 14M  & 0.3 & 16  \\
	\bottomrule
\end{tabular}%
\caption{Extrapolation results on RealEstate10k~\citep{stereo_mag}.
	All variants without \checkmark are trained with 288p, but evaluated on 576p.
	\mbox{Ours-$N$} denotes a network with $N$ PSV planes, while \textit{NoSM} refers to
	ablations without the Soft-Masking network.
} \label{tab:parameters}
\label{tab:extrapolation}
\end{table}

\paragraph{Results}
As shown in \cref{tab:extrapolation}, we outperform current state-of-the-art methods on RealEstate10k by significant margins in all metrics.
Reducing the number of depth layers decreases the performance only slightly (Ours-19).
To evaluate the Soft-Masking network, we also evaluate ablations without it (NoSM). 
However, removing the Soft-Masking network results in a noticeable drop in performance.
We note that NoSM variants still achieve significantly higher PSNR and LPIPS values than related methods (see \cref{tab:extrapolation}), however, they score lower (or similar) in SSIM compared to Stereo-Mag~\cite{stereo_mag} and 3D-Photo~\cite{3D_photo}.

We also note that FaDIV-Syn can be trained at half resolution and evaluated at full resolution without compromising much performance.
Interestingly, ablations without the SM network generalize less well to higher resolutions.

All our variants have less or equal PSV planes compared to other approaches based on plane sweeping (Stereo-Mag:~32, PB-MPI:~64, LLFF:~32).
Our main competitors Stereo-Mag~\cite{stereo_mag} and 3D-Photo~\cite{3D_photo} use significantly more learned parameters (17M and 119M), as shown in \cref{tab:extrapolation}.

\begin{figure}
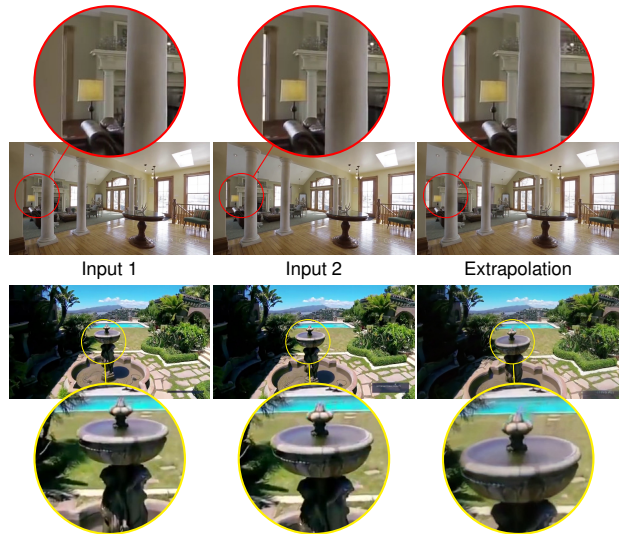
\sf \scriptsize \centering
		\begin{tikzpicture}[
	spy using outlines={circle, magnification=3.4, size=2cm, connect spies}
	]
	
	\node[inner sep=0] (img4)
	{\includegraphics[height=1.5cm]{images/qualitative_results/3054_ours19/helper_frame_6.png}};
	\spy [red] on (rel cs:x=14,y=52.5,name=img4) in node [above] at ($(img4.north)-(0,0.2)$);
	
	\node[below] at (img4.south) {Input 1};

	\node[inner sep=0,right=1pt of img4] (img5)
	{\includegraphics[height=1.5cm]{images/qualitative_results/3054_ours19/reference_frame_11.png}};
	\spy [red] on (rel cs:x=14,y=52.5,name=img5) in node [above] at ($(img5.north)-(0,0.2)$);
	
	\node[below] at (img5.south) {Input 2};
	
	\node[inner sep=0,right=1pt of img5] (img6)
	{\includegraphics[height=1.5cm]{images/qualitative_results/3054_ours19/37.png}};
	\spy [red] on (rel cs:x=14,y=52.5,name=img6) in node [above] at ($(img6.north)-(0,0.2)$);
	\node[below] at (img6.south) {Extrapolation};
	
	\end{tikzpicture}

	\begin{tikzpicture}[
	spy using outlines={circle, magnification=3.4, size=2cm, connect spies}
	]

	\node[inner sep=0] (img4)
	{\includegraphics[height=1.5cm]{images/qualitative_results/6602_ours32/helper_frame.png}};
	\spy [yellow] on (rel cs:x=47,y=50.0,name=img4) in node [below] at ($(img4.south)+(0,0.2)$);

	\node[inner sep=0,right=1pt of img4] (img5)
	{\includegraphics[height=1.5cm]{images/qualitative_results/6602_ours32/reference_frame.png}};
	\spy [yellow] on (rel cs:x=47,y=50.0,name=img5) in node [below] at ($(img5.south)+(0,0.2)$);

	\node[inner sep=0,right=1pt of img5] (img6)
	{\includegraphics[height=1.5cm]{images/qualitative_results/6602_ours32/12.png}};
	\spy [yellow] on (rel cs:x=47,y=50.0,name=img6) in node [below] at ($(img6.south)+(0,0.2)$);

	\end{tikzpicture}%
	\caption{Extrapolation on the RealEstate10k test set.}
	\label{fig:extra}%
\end{figure}

We show exemplary extrapolated views in \cref{fig:extra,fig:teaser}.

\begin{figure*}
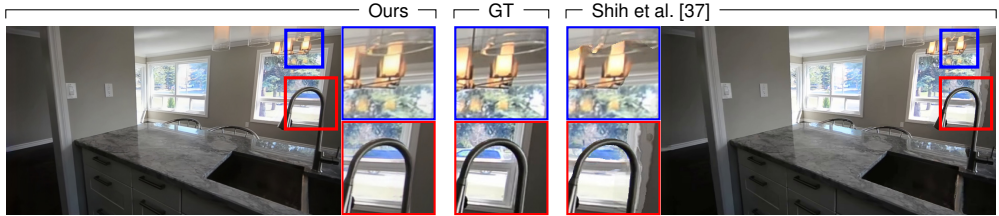
 \centering \newlength{\extraheight} \setlength{\extraheight}{2.5cm}
	\begin{tikzpicture}[font=\sffamily\scriptsize,l/.style={text depth=0pt}]
	\begin{scope}[spy using outlines={rectangle, magnification=2, size=0.49\extraheight, every spy on node/.append style={very thick}}, every spy in node/.append style={outer sep=1cm}]
	\begin{scope}[spy using outlines={rectangle, magnification=2, size=0.49\extraheight}]
	\node[inner sep=0pt] (gt) {\includegraphics[height=\extraheight]{images/qualitative_results/ours_32/2a058bafbecaccf9_65298567_65165100/2a058bafbecaccf9_65298567_65165100_zGT.png}};
	
	\spy [blue, magnification=2.5] on (rel cs:x=89,y=88,name=gt) in node [anchor=north west,outer sep=0.5pt] (bluegt) at ($(gt.north east)+(1.5,0)$);
	\spy [red, magnification=1.8] on (rel cs:x=91,y=59,name=gt) in node [anchor=north,outer sep=0.5pt] at (bluegt.south);
	\end{scope}
	
	\node[inner sep=0pt] (ours) {\includegraphics[height=\extraheight]{images/qualitative_results/ours_32/2a058bafbecaccf9_65298567_65165100/2a058bafbecaccf9_65298567_65165100_x32pred.png}};
	
	\spy [blue, magnification=2.5] on (rel cs:x=89,y=88,name=ours) in node [anchor=north west,outer sep=0.5pt] (blueours) at (ours.north east);
	\spy [red, magnification=1.8] on (rel cs:x=91,y=59,name=ours) in node [anchor=north,outer sep=0.5pt] at (blueours.south);
	
	\node[inner sep=0pt,anchor=west] (shih) at ($(ours.east)+(4.25,0)$) {\includegraphics[height=\extraheight]{images/qualitative_results/ours_32/2a058bafbecaccf9_65298567_65165100/2a058bafbecaccf9_65498767_3d_photo.png}};
	
	\spy [blue, magnification=2.5] on (rel cs:x=89,y=88,name=shih) in node [anchor=north east,outer sep=0.5pt] (blueshih) at (shih.north west);
	\spy [red, magnification=1.8] on (rel cs:x=91,y=59,name=shih) in node [anchor=north,outer sep=0.5pt] at (blueshih.south);
	\end{scope}
	
	\node [l,above=0pt of blueours] (lours) {Ours};
	\draw (lours) -| ($(ours.north west)+(0,0.1)$);
	\draw (lours) -| ($(blueours.north east)+(0,0.1)$);
	\node [l,above=0pt of bluegt] (lgt) {GT};
	\draw (lgt) -| ($(bluegt.north west)+(0,0.1)$);
	\draw (lgt) -| ($(bluegt.north east)+(0,0.1)$);
	\node [l,above=0pt of blueshih,xshift=0.5cm] (lshih) {\citet{3D_photo}};
	\draw (lshih) -| ($(blueshih.north west)+(0,0.1)$);
	\draw (lshih) -| ($(shih.north east)+(0,0.1)$);
	\end{tikzpicture}

	\caption{Extrapolation results of our method (left) compared with ground truth (center) and \citet{3D_photo} (right).}
	\label{fig:compare} \vspace{-1.5ex}
\end{figure*}

\subsection{Interpolation}\label{sec:eval:interpolation}

\begin {table}
\centering \footnotesize
\setlength{\tabcolsep}{2pt}
\begin{tabular}{l@{\hspace{8pt}}ccc@{\hspace{12pt}}r@{\hspace{-2pt}}r@{}} \toprule
	&  & $\Delta t$=5 & & \multicolumn{2}{@{}c@{}}{Depth [m]}
	\\ \cmidrule () {5-6}
	\cmidrule (r{12pt}) {2-4}
    Method & PSNR$\uparrow$  & SSIM$\uparrow$ &LPIPS$\downarrow$ &min & max \\ \midrule
	Stereo-Mag~\cite{stereo_mag}   & 28.20   & .9209    & .0783 &1.0&100  \\
	\midrule
	Ours-32   & \textbf{32.71}   & \textbf{.9448}    & \textbf{.0361} &1.0&100 \\
	Ours-32-NoSM   & 32.02   & .9343    & .0429 &1.0&100 \\
	Ours-19   & 32.17   & .9433   & .0386 &1.0&100 \\
	Ours-19-NoSM   & 31.05   & .9241    & .0464 &1.0&100 \\
	\bottomrule
\end{tabular} %
\caption{Interpolation results. Our networks are trained for 288p but evaluated in twice the resolution, i.e. 576p. We compare against Stereo-Mag~\cite{stereo_mag}, which was trained for higher resolution.}
\label{tab:eval:generalization} %
\end{table}

The second interesting problem setting is interpolation, i.e. when the target camera pose
is roughly between the two input frames of a video sequence.
For this, we randomly choose a target image $I_O$ and source frames
$I_1$ and $I_2$ before and after it, respectively. The distances $\Delta t_1,\Delta t_2$ are uniformly sampled from the interval $[4,13]$.
Note that extrapolation may still occur in this mode, since the camera seldomly moves perfectly on a straight line.
We perform both training and evaluation based on the image resolution of 518$\times$288 unless otherwise mentioned.

\begin {table}[b]
\centering \footnotesize
\setlength{\tabcolsep}{2pt}
	\begin{tabular}{l@{\hspace{8pt}}ccc@{\hspace{18pt}}ccc} \toprule
		&      & $\Delta t=5$ &        &          & $\Delta t=10$ &
		\\
		\cmidrule (r{19pt}) {2-4} \cmidrule (r{4pt}) {5-7}
		Variant &  PSNR$\uparrow$  & SSIM$\uparrow$   & LPIPS$\downarrow$ & PSNR$\uparrow$  & SSIM$\uparrow$   & LPIPS$\downarrow$\\
        \midrule
		F-32       & \textbf{33.61} & .9541          & \textbf{.0242} & 29.93          & .9205          & .0462\\
		F-32-NoSM  & 33.16          & .9493          & .0272          & 29.34          & .9071          & .0538\\
		F-19       & 33.03          & .9524          & .0263          & 29.52          & .9164          & .0497\\
		F-19-NoSM  & 32.21          & .9415          & .0293          & 28.53          & .8949          & .0579\\
		F-17-NoSM  & 32.80          & .9452          & .0304          & 28.82          & .8983          & .0606\\
		F-13-NoSM  & 31.67          & .9376          & .0326          & 28.06          & .8881          & .0634\\
		\midrule
		F-19-3view & 33.50          & \textbf{.9545} & .0248          & \textbf{29.97} & \textbf{.9215} & \textbf{.0452}\\

		\bottomrule
\end{tabular} %
\caption{Interpolation results of ablations on RealEstate10k with an image resolution of 512$\times$288 pixels. \mbox{F-13} and F-17 networks are trained with a depth distribution of $[0.3,16]$\,m, while the other networks use $[1,100]$\,m. The \textit{3view} variant uses three input views (which might not be available in certain scenarios).}
\label{tab:eval:interpolation} %
\end{table}

In \cref{tab:eval:generalization} we compare our approach to our main competitor Stereo Magnification~\cite{stereo_mag} for interpolation, which we outperform in all metrics.
Note that our networks generalize from a training image size of 512$\times$288 to 1024$\times$576 at inference, while Stereo-Mag was trained for higher resolution explicitly.
Looking at the ablations, we again achieve better generalization ability in the variants with soft masking (especially in comparison with \cref{tab:eval:interpolation}).

In our ablation study in \cref{tab:eval:interpolation}, reducing the number of PSV planes clearly reduces performance. However, removing the soft-masking network (NoSM variant) has a much bigger impact,
so that the F-19 network outperforms the F-32-NoSM variant with significantly more depth planes.

\paragraph{Three source views}\phantomsection\label{sec:eval:multiview}
In order to demonstrate FaDIV-Syn's capability to handle more than two input views, we also report the results of an F-19 variant that is trained with three input views (F-19-3view). We note that inference times and the number of parameters grow with each additional PSV plane.
In training and evaluation, we always choose a third frame outside of the two original source frames. The frame is chosen
with at least four frames distance to the other frames. \Cref{tab:eval:interpolation} shows that FaDIV-Syn benefits from more views, with the mentioned drawbacks.

\paragraph{Data Efficiency}
FaDIV-Syn generalizes well with significantly smaller training subsets (35\%, 5\%, 1\%). This experiment can be found in our supplementary material.

\subsection{Inference Time} \label{sec:timing}

\Cref{tab:timing} shows the inference time of the different models on one NVIDIA RTX 3090 GPU.
Our full model achieves 34fps on 288p input.
Removing the soft-masking network results in slightly faster inference speed.
Reducing the PSV plane count yields larger gains.
We especially note that F-19 is 43\% faster than F-32, but attains nearly the same accuracy (see \cref{tab:extrapolation}).
The fastest model is F-13-NoSM, as it only uses 13 depth planes and no SM network.

\begin {table}
\centering \footnotesize \setlength{\tabcolsep}{2.7pt}
\begin{tabular}{l@{\hspace{6pt}}rr@{\hspace{12pt}}rr@{\hspace{12pt}}rr}
	\toprule
	& \multicolumn{2}{c@{\hspace{9pt}}}{Torch [ms]}
	& \multicolumn{2}{c@{\hspace{9pt}}}{TRT-32 [ms]}
	& \multicolumn{2}{c}{TRT-16 [ms]}
	\\
	\cmidrule (r{9pt}) {2-3} \cmidrule (r{9pt}) {4-5} \cmidrule (r{3pt}) {6-7} 
	Model   & 288p & 540p & 288p & 540p & 288p & 540p   \\
	\midrule
	F-32      &  28.7         &   96.8        &   27.1       & 92.2          & 12.3  & 40.4   \\
	F-32-NoSM &  27.1         &   94.1        &  25.9        & 89.2          & 11.3  & 39.7    \\
	F-19      &  16.3         &   55.2        &  14.7        & 49.3          & 7.7  & 25.3  \\
	F-19-NoSM &  15.1         &   51.5        &  12.1        & 39.8          & 5.2  & 19.3  \\
	F-17-NoSM &  13.5         &   46.3        & 11.2         & 35.8          & 4.8  & 17.4 \\
	F-13-NoSM & \textbf{10.7} & \textbf{35.8} & \textbf{9.0} & \textbf{29.7} & \textbf{3.9}  & \textbf{14.0} \\
	\bottomrule

\end{tabular} 
\caption{Inference times of our networks on RTX 3090.
	We show native PyTorch and TensorRT (TRT)~\citep{NVIDIA-TensorRT} float32/float16
	versions.
	The times do not include PSV generation (1.5\,ms @ 540p).}
\label{tab:timing} %
\end{table}

In addition to results on vanilla PyTorch, \cref{tab:timing} also shows inference times in TensorRT \cite{NVIDIA-TensorRT},
which already boosts performance for float32 precision without losing accuracy.
TensorRT offers possibilities to quantize the weights of neural networks to float16. This quantization of our models retains almost 100\% accuracy for SSIM and PSNR and approximately 99\% for LPIPS.
Further, it leads to a significant acceleration: We can achieve up to 71\,fps on 960$\times$540 images and 256\,fps for 512$\times$288 resolution.
In comparison, 3D-Photo and Stereo-Mag process 2-3\,min and 93\,ms per 540p image, respectively.

If faster processing times are desired without removing the soft-masking network, one can also estimate the soft-masks in half resolution and upsample them without losing much accuracy (see \cref{tab:multiscale}).

\begin {table}[b] \centering \footnotesize  \setlength{\tabcolsep}{2pt}

\begin{tabular}{@{}l@{\hspace{4pt}}ccc@{\hspace{10pt}}ccc} \toprule
	& \multicolumn{3}{c@{\hspace{10pt}}}{Accuracy} &  \multicolumn{3}{c}{540p [ms]}  \\
	\cmidrule (r{9pt}) {2-4} \cmidrule (r{3pt}) {5-7}
	Method     & SSIM$\uparrow$    & PSNR$\uparrow$    & LPIPS$\downarrow$ & Torch & TRT-32 & TRT-16 \\
	\midrule
	F-32                       & \textbf{.9020}  & \textbf{29.39}   & \textbf{.0556}       & 96.8        & 89.2 & 39.7\\
	F-32 (mixed scale)                        & .8967           & 28.99   & .0623          & 76.5  & 71.9 & 32.8       \\
	F-19                      & .8985           & 29.04   & .0583          & 55.2   & 49.3  &  25.3    \\
    F-19 (mixed scale)                       & .8943           & 28.74   & .0611          & \textbf{43.4} & \textbf{37.4} & \textbf{19.4} \\

	\bottomrule
\end{tabular}%
\caption{Mixed scale inference with 288p mask generation and 576p fusion as described in \cref{sec:timing}. We show extrapolation performance and timings.} \label{tab:multiscale}
\end{table}

\subsection{Qualitative Results}

\paragraph{Continuous Depth}
The fixed plane depths do not constrain FaDIV-Syn. This effect can particularly be seen
on straight lines across different depths, which are preserved by our approach (see \cref{fig:view_extrapolation_scene} and suppl. material).
We conclude that FaDIV-Syn does not directly propagate information from the PSV, but properly interpolates
between planes.

\paragraph{Occlusions \& Disocclusions}

\begin{figure}
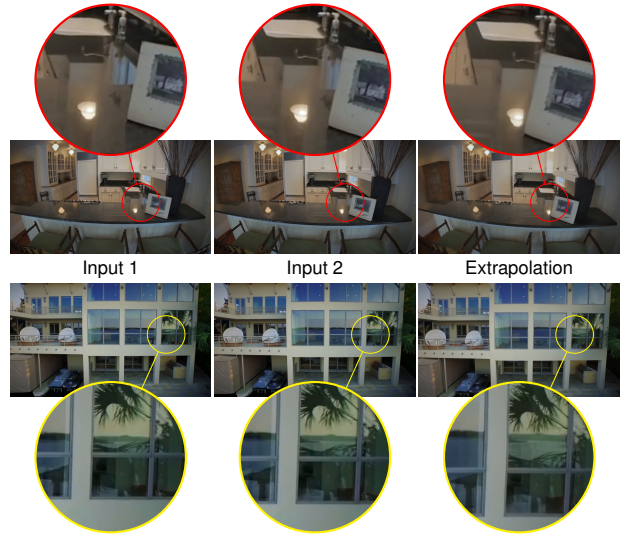
\sf \scriptsize \centering
	\vspace{-0.5ex}
	\begin{tikzpicture}[
	spy using outlines={circle, magnification=4, size=2cm, connect spies}
	]
	
	\node[inner sep=0] (img1)
	{\includegraphics[height=1.5cm]{images/qualitative_results/496_ours32/helper_frame.png}};
	\spy [red] on (rel cs:x=65,y=45,name=img1) in node [above] at ($(img1.north)+(0,-0.2)$);
	
	\node[below] at (img1.south) {Input 1};
	
	\node[inner sep=0,right=1pt of img1] (img2)
	{\includegraphics[height=1.5cm]{images/qualitative_results/496_ours32/reference_frame.png}};
	\spy [red] on (rel cs:x=65,y=45,name=img2) in node [above] at ($(img2.north)+(0,-0.2)$);
		
	\node[below] at (img2.south) {Input 2};
	
	\node[inner sep=0,right=1pt of img2] (img3)
	{\includegraphics[height=1.5cm]{images/qualitative_results/496_ours32/19.png}};
	\spy [red] on (rel cs:x=65,y=45,name=img3) in node [above] at ($(img3.north)+(0,-0.2)$);
	
	\node[below] at (img3.south) {Extrapolation};
	\end{tikzpicture}

	\begin{tikzpicture}[
	spy using outlines={circle, magnification=4, size=2cm, connect spies}
	]
	
	\node[inner sep=0] (img1)
	{\includegraphics[height=1.5cm]{images/qualitative_results/3407_ours32/helper_frame.png}};
	\spy [yellow] on (rel cs:x=77.5,y=55,name=img1) in node [below] at ($(img1.south)+(0,+0.2)$);

	\node[inner sep=0,right=1pt of img1] (img2)
	{\includegraphics[height=1.5cm]{images/qualitative_results/3407_ours32/reference_frame.png}};
	\spy [yellow] on (rel cs:x=77.5,y=55,name=img2) in node [below] at ($(img2.south)+(0,+0.2)$);

	\node[inner sep=0,right=1pt of img2] (img3)
	{\includegraphics[height=1.5cm]{images/qualitative_results/3407_ours32/19.png}};
	\spy [yellow] on (rel cs:x=77.5,y=55,name=img3) in node [below] at ($(img3.south)+(0,+0.2)$);
	
	\end{tikzpicture}
\caption{FaDIV-Syn is able to represent multiple layers of depth at one location and thus handles reflections correctly.}
\label{fig:reflections} \vspace{-2ex}
\end{figure}

\cref{fig:teaser,fig:view_extrapolation_scene,fig:extra} show examples of disocclusions. Our method handles disocclusions for both interpolation and extrapolation.
As shown in \cref{fig:teaser,fig:reflections}, FaDIV-Syn can also handle and represent occlusions.

\paragraph{Reflections and Transparency}

Often methods that use depth have problems in representing transparencies and reflections (see suppl. material), since there is often more than one depth value at a certain pixel location. FaDIV-Syn is designed in such a way that there is not only one depth for each pixel, but a multitude of information in the different depth layers. This allows recognition of the correct position of both the surface and the reflection on it, as shown in \cref{fig:reflections}. Examples for transparencies can be found in \cref{fig:view_extrapolation_scene,fig:compare}.

\begin{figure}
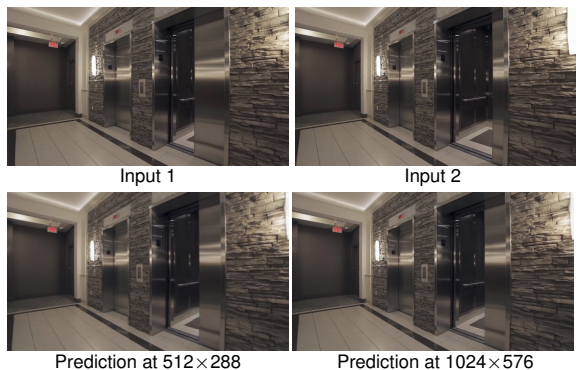
 \centering \scriptsize \sf \setlength{\tabcolsep}{0.75pt} \setlength{\imgheight}{2.1cm}
\begin{tabular}{ccc}
\includegraphics[height=\imgheight]{images/qualitative_results/127_elevator/helper_frame.png}
& \includegraphics[height=\imgheight]{images/qualitative_results/127_elevator/reference_frame.png}\vspace*{-.5mm}\\
Input 1 & Input 2 \vspace*{.5mm}\\
\includegraphics[height=\imgheight]{images/qualitative_results/127_elevator/13_288.png} 

& \includegraphics[height=\imgheight]{images/qualitative_results/127_elevator/13_576.png} \vspace*{-.5mm}\\
Prediction at 512$\times$288
& Prediction at 1024$\times$576\vspace*{-1mm}
\end{tabular}
\caption{Interpolation from two reference views with a moving object (elevator door). See supplementary video for animation.
This figure has been created with a 17-NoSM network.}
\label{fig:elevator} \vspace{-1ex}
\end{figure}

\paragraph{Moving Objects}
While dynamic scenes are more difficult for
view synthesis (and one could argue that the problem is ill-posed), FaDIV-Syn nonetheless shows
plausible behavior here.
For example, it has learned to interpolate between positions of movable objects (see \cref{fig:elevator}).
This would be hard to achieve if a method estimates depth first or only blends pixels.

\subsection{Limitations}

Our approach sometimes struggles with inpainting under large camera movements. One example is the fountain/pool boundary in \cref{fig:extra}.
We believe that adversarial losses could help to encourage realistic inpainting.
Furthermore, we expect that semi-supervised techniques such as proposed by~\citet{hani2020continuous} could be used
to increase the robustness of the method against arbitrary target poses, since the supervision offered by RealEstate10k
only covers inter- and extrapolation on smooth camera trajectories.
Additionally, the PSV depth distribution is currently fixed and techniques such as depth plane resampling~\citep{nguyen2020rgbd} could be advantageous for varying geometries.
Another issue we observed is that our network shows very small learning progress after several epochs. While we stop the training after a fixed number of epochs, the validation and testing score still improves for a long time, indicating
that further gains are possible through changes in the training regime.
Finally, the inference time is limited by the network itself, where compression techniques~\citep{Luo_2017_ICCV} could be applied to reduce network
runtime even further.
Note that additional demonstrations of failure cases can be found in the supplementary material.

\section{Conclusion}

We introduced FaDIV-Syn, a fast depth-independent novel view synthesis method that exceeds state-of-the-art performance in interpolation and extrapolation on the RealEstate10k
dataset. The method generalizes well to larger resolutions.
Furthermore, its lightweight architecture makes our method real-time-capable with 25-256\,fps, depending
on output resolution and desired quality. The fast inference times make it applicable for live applications.
The proposed gating module encourages self-supervised learning of soft masks, which noticeably
improves performance and provides valuable insight into the network operation.
Overall, we conclude that the direct usage of the PSV for RGB view synthesis is a promising approach
especially for real-time applications and will inspire further research in this direction.

\section*{Acknowledgments}
This work was funded by grant BE 2556/16-2 (Research Unit FOR 2535 Anticipating Human Behavior) of the German Research Foundation (DFG) and the Federal Ministry of Education and Research of Germany as part of the competence center for machine learning ML2R (01IS18038C).

{\small
\bibliography{references}
}
\ifdefined\includesupp
\pagebreak
\clearpage
\newpage
\appendices

\section{Network Architecture Details}

Here, we provide more details regarding the architectures of our Soft-Masking network (see \cref{tab:sm:architecture}), our fusion network (see \cref{tab:fusion:architecture}) as well as the NoSM network architecture (see \cref{tab:nosm:architecture}).

\begin {table}[h!]
\centering \footnotesize \setlength{\tabcolsep}{5pt}
\begin{tabular}{crrc} \toprule
	{Input}           & {$k$} & {$c$}  & Output \\ \midrule
	$\mathrm{gray}(PSV)$                 & 3      & $\min(2 \cdot N, 256)$ & $down_1$\\
	\textbf{Pool}$(down_1)$ &              3      & $\min(2^3 \cdot N, 256)$&  $down_2$\\
	\textbf{Pool}$(down_2)$ &              3      & $\min(2^4 \cdot N, 256)$&  $down_3$\\
	\textbf{Pool}$(down_3)$ &              3      & $\min(2^5 \cdot N, 256)$&  $down_4$\\

	\textbf{Pool}$(down_4)$              & 3      & $\min(2^4 \cdot N, 256)$&  $up_1$\\
	\textbf{Up}$(up_1), down_3$           & 3      & $\min(2^3 \cdot N, 256)$&  $up_2$\\
	\textbf{Up}$(up_2), down_2$           & 3      & $\min(2^2 \cdot N, 256)$&  $up_3$\\
	\textbf{Up}$(up_3), down_1$           & 3      & $\min(2^1 \cdot N, 256)$&  $up_4$\\
	\textbf{Up}$(up_4), P$& 3   & {$2\cdot N$}&  $final$\\
	$\mathrm{softmax}(masking)$                   & -   & $N$&  $pred$\\ \bottomrule
\end{tabular}
\caption{Soft-Masking network architecture for $N$ grayscale depth planes in the PSV.
Each row denotes a convolutional layer, where $k$ is the kernel size and $c$ is the number of
output features. \textbf{Pool} is $2 \times 2$ average pooling, and \textbf{Up} denotes bilinear upsampling with a factor of 2.}
\label{tab:sm:architecture}
\end{table}

\begin {table}[h!]
\centering \footnotesize \setlength{\tabcolsep}{4pt}
\begin{tabular}{crrrrc} \toprule
	{Input}           & {$k_1$} & {$c_1$}  & {$k_2$} & {$c_2$} & Output \\ \midrule
	\textbf{$PSV$}        & 3      & {$6\cdot N+N$}& 3    & {$12 \cdot N$}   & $groupconv$\\
	\textbf{G}$(groupconv)$        & 3      & {$6\cdot N$}& 3    & {$3 \cdot N$}   & $bottleneck$\\
	$bottleneck$ & 3      & {$6\cdot N$}& 3    & {$6 \cdot N$}   & $down_1$\\
	$down_1$ & 3      & {$128$}& 3    & {$128$}   & $down_2$\\
	$down_2$ & 3      & {$256$}& 3    & {$256$}   & $down_3$\\
	$down_3$ & 3      & {$256$}& 3    & {$256$}  & $down_4$\\
	$down_4$ & 3      & {$256$}& 3    & {$256$}   & $dilated$\\
	\textbf{Up}$(dilated),down_3$& 3   & {$256$}& 3    & {$128$}   & $up_1$\\
	\textbf{Up}$(up_1),down_2$& 3   & {$128$}& 3    & {$6 \cdot N$}   & $up_2$\\
	\textbf{Up}$(up_2),down_1$& 3   & {$6\cdot N$}& 3    & {$3 \cdot N$}   & $up_3$\\
	\textbf{Up}$(up_3),bottleneck$& 3   & {$32$}& 3    & {$32$}   & $up_4$\\
	$up_4$                   & 1   & {$3$}& {$-$}    & {$-$}   & $pred$\\ \bottomrule
\end{tabular}
\caption{Fusion network with N depth planes in the PSV. Each row shows 2 convolutional layers, where $k$ is the kernel size and $c$ is the number of output features.
	\textbf{G} denotes the gating operation and \textbf{Up} denotes upsampling.}
\label{tab:fusion:architecture}
\end{table}

\begin {table}[h!]
\centering \footnotesize \setlength{\tabcolsep}{4pt}
\begin{tabular}{crrrrc} \toprule
	{Input}           & {$k_1$} & {$c_1$}  & {$k_2$} & {$c_2$} & Output \\ \midrule
   \textbf{$PSV$}         & 3      & {$6\cdot N$}& 3    & {$12 \cdot 6$}   & $groupconv$\\
   \textbf{G}$(groupconv)$        & 3      & {$6 \cdot N$}& 3    & {$6 \cdot N$}   & $bottleneck$\\
   \textbf{G}$(bottleneck)$ & 3      & {$6 \cdot N$}& 3    & {$12 \cdot N$}   & $down_1$\\
   \textbf{G}$(down_1)$ & 3      & {$128$}& 3    & {$256$}   & $down_2$\\
   \textbf{G}$(down_2)$ & 3      & {$256$}& 3    & {$512$}   & $down_3$\\
   \textbf{G}$(down_3)$ & 3      & {$256$}& 3    & {$1024$}  & $down_4$\\
   \textbf{G}$(down_4)$ & 3      & {$512$}& 3    & {$512$}   & $dilated$\\
   \textbf{Up}$(dilated),G(down_3)$& 3   & {$256$}& 3    & {$256$}   & $up_1$\\
   \textbf{Up}$(up_1),G(down_2)$& 3   & {$12 \cdot N$}& 3    & {$12 \cdot N$}   & $up_2$\\
   \textbf{Up}$(up_2),G(down_1)$& 3   & {$6 \cdot N$}& 3    & {$6 \cdot N$}   & $up_3$\\
   \textbf{Up}$(up_3),G(bottleneck)$& 3   & {$32$}& 3    & {$32$}   & $up_4$\\
   $up_4$                   & 1   & {$3$}& {$-$}    & {$-$}   & $pred$\\ \bottomrule
\end{tabular}
\caption{NoSM architecture with N depth planes in the PSV. Each row shows 2 convolutional layers, where $k$ is the kernel size and $c$ is the number of output features.
\textbf{G} denotes the gating operation which reduces the number of feature maps by factor 2 and \textbf{Up} denotes upsampling.}
\label{tab:nosm:architecture}
\end{table}

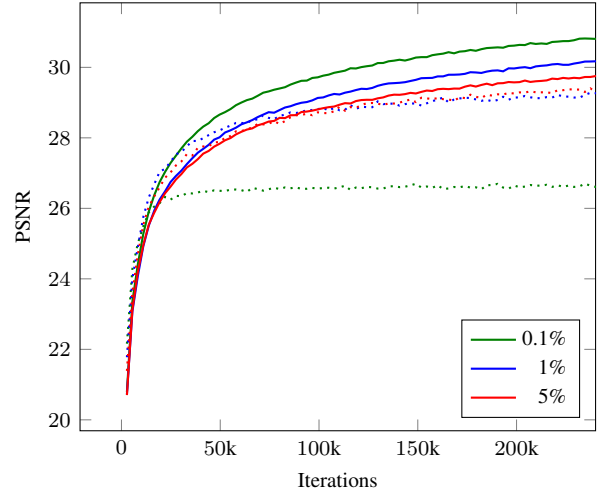
\begin{figure}[h!] \centering \vspace{-1ex}
 \begin{tikzpicture}[font=\footnotesize]
  \begin{axis}[
        xmax=240000,
        xlabel={Iterations},
        ylabel={PSNR},
        legend pos=south east,
        legend cell align={right},
        cycle list/Set1-3,
        cycle multi list={
            green!50!black,blue,red\nextlist
            solid,dotted\nextlist
        },
        every axis plot/.style={
            no marks,
            thick
        },
        change x base,
        x SI prefix=kilo,
        xticklabel={
            \ifdim\tick pt=0pt
                $\pgfmathprintnumber{\tick}$
            \else
                $\pgfmathprintnumber{\tick}$k
            \fi
        },
    ]
    \addplot table [x expr={\thisrow{iteration}*4/5},y=psnr] {train_progress/train_log_0_1perc.txt};
    \addplot table [x expr={\thisrow{iteration}*4/5},y=psnr] {train_progress/val_log_0_1perc.txt};

    \addplot table [x expr={\thisrow{iteration}*4/5},y=psnr] {train_progress/train_log_1perc.txt};
    \addplot table [x expr={\thisrow{iteration}*4/5},y=psnr] {train_progress/val_log_1perc.txt};

    \addplot table [x expr={\thisrow{iteration}*4/5},y=psnr] {train_progress/train_log_5perc.txt};
    \addplot table [x expr={\thisrow{iteration}*4/5},y=psnr] {train_progress/val_log_5perc.txt};

    \legend{{0.1\%,,1\%,,5\%}};
  \end{axis}
 \end{tikzpicture} \vspace{-1.5ex}
 \caption{PSNR on reduced training (solid) and validation (dotted) splits of the RealEstate10k dataset during 17-NoSM network training.}
 \label{fig:training_curve}
\end{figure}

\begin{figure} [h!] \centering
 \begin{tikzpicture}[font=\footnotesize]
  \begin{axis}[
        xmax=240000,
        xlabel={Iterations},
        ylabel={Perceptual loss},
        legend pos=north east,
        legend cell align={right},
        cycle list/Set1-3,
        cycle multi list={
            green!50!black,blue,red\nextlist
            solid,dotted\nextlist
        },
        every axis plot/.style={
            no marks,
            thick
        },
        change x base,
        x SI prefix=kilo,
        xticklabel={
            \ifdim\tick pt=0pt
                $\pgfmathprintnumber{\tick}$
            \else
                $\pgfmathprintnumber{\tick}$k
            \fi
        },
        width=\linewidth,
        height=.8\linewidth
    ]
    \addplot table [x expr={\thisrow{iteration}*4/5},y=percept] {train_progress/train_log_0_1perc.txt};
    \addplot table [x expr={\thisrow{iteration}*4/5},y=percept] {train_progress/val_log_0_1perc.txt};

    \addplot table [x expr={\thisrow{iteration}*4/5},y=percept] {train_progress/train_log_1perc.txt};
    \addplot table [x expr={\thisrow{iteration}*4/5},y=percept] {train_progress/val_log_1perc.txt};

    \addplot table [x expr={\thisrow{iteration}*4/5},y=percept] {train_progress/train_log_5perc.txt};
    \addplot table [x expr={\thisrow{iteration}*4/5},y=percept] {train_progress/val_log_5perc.txt};

    \legend{{0.1\%,,1\%,,5\%}};
  \end{axis}
 \end{tikzpicture}
 \caption{Perceptual loss on reduced training (solid) and validation (dotted) splits of the RealEstate10k dataset during 17-NoSM network training.}
 \vspace{0.3cm}
 \label{fig:training_curve_lpips}
\end{figure}
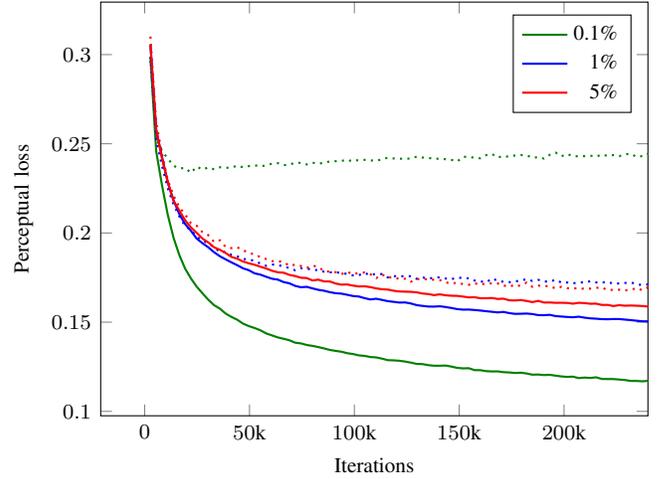

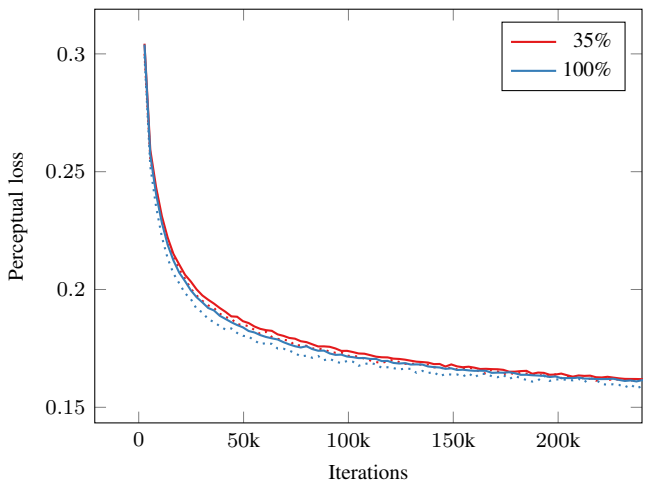
\begin{figure} [b!] \centering
 \begin{tikzpicture}[font=\footnotesize]
  \begin{axis}[
        xmax=240000,
        xlabel={Iterations},
        ylabel={Perceptual loss},
        legend pos=north east,
        legend cell align={right},
        cycle list/Set1-4,
        cycle multi list={
            Set1-4\nextlist
            solid,dotted\nextlist
        },
        every axis plot/.style={
            no marks,
            thick
        },
        change x base,
        x SI prefix=kilo,
        xticklabel={
            \ifdim\tick pt=0pt
                $\pgfmathprintnumber{\tick}$
            \else
                $\pgfmathprintnumber{\tick}$k
            \fi
        },
        width=\linewidth,
        height=.8\linewidth
    ]

    \addplot table [x expr={\thisrow{iteration}*4/5},y=percept] {train_progress/train_log_35perc.txt};
    \addplot table [x expr={\thisrow{iteration}*4/5},y=percept] {train_progress/val_log_35perc.txt};

    \addplot table [x expr={\thisrow{iteration}*4/5},y=percept] {train_progress/train_log_100perc.txt};
    \addplot table [x expr={\thisrow{iteration}*4/5},y=percept] {train_progress/val_log_100perc.txt};

    \legend{{35\%,,100\%}};
  \end{axis}
 \end{tikzpicture}
 \caption{Perceptual loss on reduced training (solid) and validation (dotted) splits of the RealEstate10k dataset during 17-NoSM network training.}
 \label{fig:training_curve_35_lpips}
\end{figure}

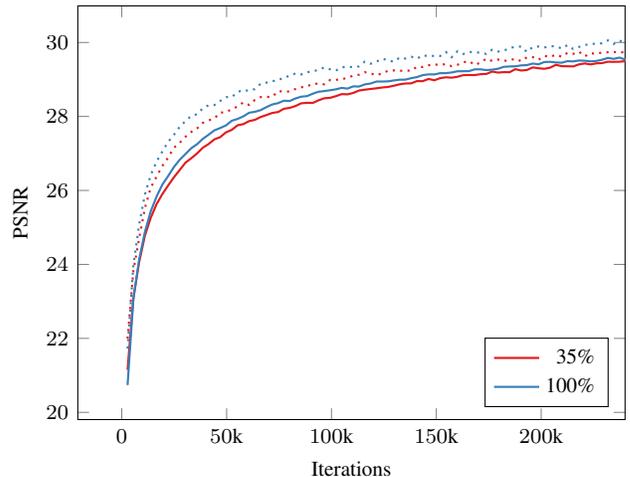
\begin{figure} \centering
 \begin{tikzpicture}[font=\footnotesize]
  \begin{axis}[
        xmax=240000,
        xlabel={Iterations},
        ylabel={PSNR},
        legend pos=south east,
        legend cell align={right},
        cycle list/Set1-4,
        cycle multi list={
            Set1-4\nextlist
            solid,dotted\nextlist
        },
        every axis plot/.style={
            no marks,
            thick
        },
        change x base,
        x SI prefix=kilo,
        xticklabel={
            \ifdim\tick pt=0pt
                $\pgfmathprintnumber{\tick}$
            \else
                $\pgfmathprintnumber{\tick}$k
            \fi
        },
        width=\linewidth,
        height=.8\linewidth
    ]

    \addplot table [x expr={\thisrow{iteration}*4/5},y=psnr] {train_progress/train_log_35perc.txt};
    \addplot table [x expr={\thisrow{iteration}*4/5},y=psnr] {train_progress/val_log_35perc.txt};

    \addplot table [x expr={\thisrow{iteration}*4/5},y=psnr] {train_progress/train_log_100perc.txt};
    \addplot table [x expr={\thisrow{iteration}*4/5},y=psnr] {train_progress/val_log_100perc.txt};

    \legend{{35\%,,100\%}};
  \end{axis}
 \end{tikzpicture}
 \caption{PSNR on reduced training (solid) and validation (dotted) splits of the RealEstate10k dataset during 17-NoSM network training.} \vspace{-1ex}
 \label{fig:training_curve_35}
\end{figure}

\begin{table*}[h!] \center \scriptsize
\setlength{\tabcolsep}{2.5pt}
\resizebox{\linewidth}{!}{%
	\begin{tabular}{l@{\hspace{4pt}}l@{\hspace{17pt}}ccc@{\hspace{17pt}}ccc@{\hspace{17pt}}ccc} \toprule
			\parbox[t]{2mm}{\multirow{2}{*}[-4pt]{\rotatebox[origin=c]{90}{Train}}} & &  & $\Delta t=2$ &        &     & $\Delta t=5$ &        &          & $\Delta t=10$ &
			\\
			\cmidrule (r{14pt}) {3-5} \cmidrule (r{14pt}) {6-8} \cmidrule (r{0pt}) {9-11}
			& Model & PSNR $\uparrow$ & SSIM $\uparrow$ &LPIPS $\downarrow$& PSNR $\uparrow$ & SSIM $\uparrow$ & LPIPS $\downarrow$ & PSNR $\uparrow$ & SSIM $\uparrow$ & LPIPS $\downarrow$\\ \midrule
			\parbox[t]{2mm}{{\rotatebox[origin=c]{90}{\textbf{0.1\%}}}}
			& 17-NoSM & 31.91$_{-\text{11.7\%}}$   & .9361$_{-\text{3.19\%}}$   & .0381$_{\text{+140\%}}$ & 28.03$_{-\text{13.5\%}}$  & .8825$_{-\text{6.32\%}}$   & .0712$_{\text{+122\%}}$   & 24.70$_{-\text{13.4\%}}$  & .8129$_{-\text{9.07\%}}$   & .1257$_{\text{+103\%}}$\\ \\
			\parbox[t]{2mm}{{\rotatebox[origin=c]{90}{\textbf{1\%}}}}
			& 17-NoSM & 34.93$_{-\text{3.31\%}}$   & .9607$_{-\text{0.64\%}}$    & .0191$_{\text{+20.1\%}}$ & 31.38$_{-\text{3.20\%}}$  & .9320$_{-\text{1.06\%}}$   & .0360$_{\text{+12.2\%}}$   & 27.60$_{-\text{3.24\%}}$  & .8789$_{-\text{1.69\%}}$   & .0695$_{\text{+12.1\%}}$ \\ \\
			\parbox[t]{2mm}{{\rotatebox[origin=c]{90}{\textbf{5\%}}}}
			& 17-NoSM & 35.19$_{-\text{2.58\%}}$    & .9616$_{-\text{0.55\%}}$    & .0173$_{\text{+8.81\%}}$  & 31.92$_{-\text{1.54\%}}$  & .9366$_{-\text{0.57\%}}$   & .0327$_{\text{+1.87\%}}$   & 28.20$_{-\text{1.16\%}}$  & .8874$_{-\text{0.74\%}}$   & .0631$_{\text{+1.77\%}}$\\ \\
			\parbox[t]{2mm}{{\rotatebox[origin=c]{90}{\textbf{35\%}}}}
			& 17-NoSM &\textbf{36.15}$_{\text{+0.08\%}}$   & .9658$_{-\text{0.11\%}}$    & \textbf{.0158}$_{-\text{0.63\%}}$  & \textbf{32.52}$_{\text{+0.32\%}}$   & .9417$_{-\text{0.03\%}}$  & \textbf{.0307}$_{-\text{4.36\%}}$    & \textbf{28.59}$_{\text{+0.24\%}}$  & \textbf{.8944}$_{\text{+0.05\%}}$    & \textbf{.0605}$_{-\text{2.42\%}}$ \\ \\
			\parbox[t]{2mm}{{\rotatebox[origin=c]{90}{\textbf{100\%}}}}
			& 17-NoSM & 36.12$_{\text{+0.00\%}}$   & \textbf{.9669}$_{\text{+0.00\%}}$   & .0159$_{\text{+0.00\%}}$  & 32.42$_{\text{+0.00\%}}$  & \textbf{.9420}$_\text{{+0.00\%}}$   & .0321$_{\text{+0.00\%}}$   & 28.53$_{\text{+0.00\%}}$   & .8940$_{\text{+0.00\%}}$   & .0620$_{\text{+0.00\%}}$ \\
			\bottomrule
	\end{tabular}} \vspace{-1.5ex}
	\caption{Data efficiency experiment. The \textit{Train} column shows the training dataset
		size relative to the full RealEstate10k train split.} \vspace{-2.5ex}
	\label{tab:eval:data}
\end{table*}

\section{Extended Data Efficiency Results}

Due to lack of space in the main paper, we shifted details on the data efficiency experiments that belong to Section 4.2 into the supplemental.
We train our 17-NoSM ablation on smaller fractions of the full RealEstate10k training dataset, and evaluate on the full test set.
The dataset size is reduced by randomly choosing scenes until the specified size is met (35\%, 5\%, 1\%, and 0.1\%).
As shown in \cref{tab:eval:data}, all sizes from 35\% to 1\% give sufficiently good results, where 35\% even performs similarly or slightly better than our model trained on the full dataset.
It is possible that further training may yield advantages, since we set an upper bound on the training iterations as described in Section 3.3.
However, we conclude that 35\% of RealEstate10k still contains enough scene and pose variance to prevent the network from overfitting
(see \cref{fig:training_curve_35}).
This is to be expected, since the triplet sampling during training greatly augments the number of training samples.
The 1\% network maintains good performance for SSIM and PSNR but starts losing significantly in LPIPS.
Finally, the 0.1\% network loses significant performance in all metrics and seems to be outside of the boundary for satisfactory results.
As \cref{fig:training_curve,fig:training_curve_lpips} show, we observed significant drops in validation performance for the 1\% and 0.1\% training split.
Starting with 35\%, we observe that the validation score is actually better than the training score (see \cref{fig:training_curve_35,fig:training_curve_35_lpips}),
which is caused by batch normalization:
The average parameters used during evaluation seem to work more robustly than the on-line statistics computed for each batch
during training.
Overall, we conclude that above 35\% there are no indications of overfitting at all.

\section{Limitations and Failure Cases}
We present some examples for failure cases in \cref{fig:inpainting,fig:pose,fig:fail}.

\paragraph{Inpainting}
Even though our method is designed in such a way that inpainting and PSV fusion can be performed simultaneously it often struggles to inpaint large missing regions. We visualize two examples in \cref{fig:inpainting}. We expect that this may be a result of (1) the limited receptive field and (2) a number of learned parameters which is insufficient for filling in large regions with reasonable and realistic content. We also show 3D-Photo~\cite{3D_photo} results in \cref{fig:inpainting}, which uses a separate inpainting network. However, we believe that both our method and 3D-Photo perform similar in inpainting regions, so this does not seem to be an architectural advantage.

\paragraph{Camera pose errors}
The RealEstate10k dataset has been annotated with ORB-SLAM2~\citep{mur2017orb} and bundle adjustment. This sometimes leads to inaccurate camera poses. While our method can generally handle small misalignments, larger errors can cause blurred regions as demonstrated in \cref{fig:pose}. We expect that it could be advantageous to allow small camera pose corrections instead of assuming that they are fixed. However, predicting camera offsets must be embedded into the pipeline in a learned fashion, unless per scene optimization is desired.

\begin {table} \centering \footnotesize  %
\begin{tabular}{lrrrr} \toprule

	Method        & Iterations & SSIM$\uparrow$    & PSNR$\uparrow$    & LPIPS$\downarrow$ \\
	\midrule
	Ours-19       & 330k                        & .8985           & 29.04            & .0583    \\
	Ours-19       & 660k                        & .8989           & \textbf{29.27}            & .0558    \\
	Ours-19       & 990k                        & \textbf{.9008}           & 29.19            & \textbf {.0552}    \\

	\bottomrule
\end{tabular}
\vspace{-1ex}
\caption{Long-time training behavior. The table shows extrapolation results on the RealEstate10k~\citep{stereo_mag} test set.
	All variants are trained with 288p, but evaluated on 576p.
} \label{tab:parameters}
\vspace{-1ex}
\label{tab:more_train}
\end{table}

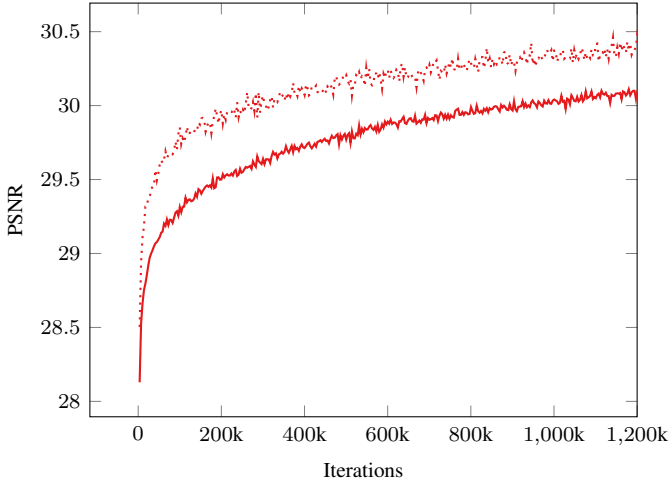
\begin{figure}[h!] \centering
\begin{tikzpicture}[font=\footnotesize]
\begin{axis}[
xmax=1200000,
xlabel={Iterations},
ylabel={PSNR},
cycle list/Set1-4,
cycle multi list={
	Set1-4\nextlist
	solid,dotted\nextlist
},
every axis plot/.style={
	no marks,
	thick
},
change x base,
x SI prefix=kilo,
xticklabel={
	\ifdim\tick pt=0pt
	$\pgfmathprintnumber{\tick}$
	\else
	$\pgfmathprintnumber{\tick}$k
	\fi
},
width=\linewidth,
height=.8\linewidth
]

\addplot table [x=iteration,y=psnr] {train_progress/train_log_corr19_extra.txt};
\addplot table [x=iteration,y=psnr] {train_progress/val_log_corr19_extra.txt};

\end{axis}
\end{tikzpicture}
\caption{Long-time training behavior. We show training (solid) and validation (dotted) PSNR during training
for more iterations.} \vspace{-1ex}
\label{fig:training_curve_corr19}
\end{figure}

\begin{figure*}
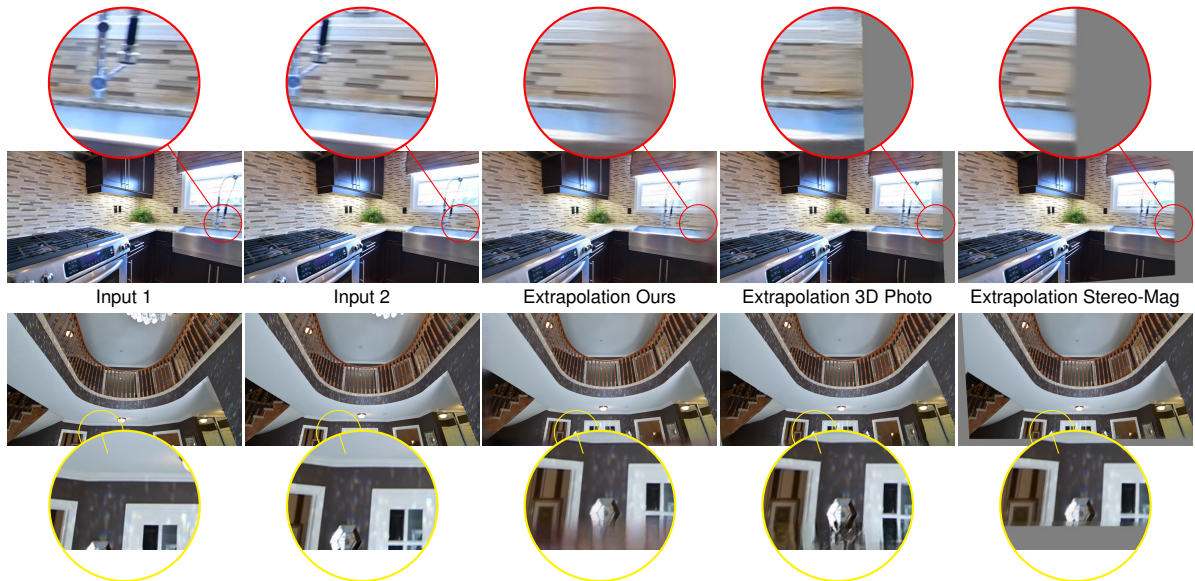
\sf \scriptsize \centering
	\begin{tikzpicture}[
	spy using outlines={circle, magnification=4, size=2cm, connect spies}
	]
	
	\node[inner sep=0] (img1)
	{\includegraphics[height=1.75cm]{images/failure_cases/scene_1/source.png}};
	\spy [red] on (rel cs:x=92,y=45,name=img1) in node [above] at ($(img1.north)+(0,-0.1)$);
	
	\node[below] at (img1.south) {Input 1};
	
	\node[inner sep=0,right=1pt of img1] (img2)
	{\includegraphics[height=1.75cm]{images/failure_cases/scene_1/ref.png}};
	\spy [red] on (rel cs:x=92,y=45,name=img2) in node [above] at ($(img2.north)+(0,-0.1)$);
	
	\node[below] at (img2.south) {Input 2};
	
	\node[inner sep=0,right=1pt of img2] (img3)
	{\includegraphics[height=1.75cm]{images/failure_cases/scene_1/ours32.png}};
	\spy [red] on (rel cs:x=92,y=45,name=img3) in node [above] at ($(img3.north)+(0,-0.1)$);
	
	\node[below] at (img3.south) {Extrapolation Ours};
	
	\node[inner sep=0,right=1pt of img3] (img4)
	{\includegraphics[height=1.75cm]{images/failure_cases/scene_1/3d_photo.png}};
	\spy [red] on (rel cs:x=92,y=45,name=img4) in node [above] at ($(img4.north)+(0,-0.1)$);
	
	\node[below] at (img4.south) {Extrapolation 3D Photo};
	
	\node[inner sep=0,right=1pt of img4] (img5)
	{\includegraphics[height=1.75cm]{images/failure_cases/scene_1/ostereo_mag.png}};
	\spy [red] on (rel cs:x=92,y=45,name=img5) in node [above] at ($(img5.north)+(0,-0.1)$);
	
	\node[below] at (img5.south) {Extrapolation Stereo-Mag};
	\end{tikzpicture}
	
	\begin{tikzpicture}[
	spy using outlines={circle, magnification=3.4, size=2cm, connect spies}
	]
	
	\node[inner sep=0] (img6)
	{\includegraphics[height=1.75cm]{images/failure_cases/0ef68/source.png}};
	\spy [yellow] on (rel cs:x=40,y=10.0,name=img1) in node [below] at ($(img6.south)+(0,0.2)$);

	\node[inner sep=0,right=1pt of img1] (img7)
	{\includegraphics[height=1.75cm]{images/failure_cases/0ef68/ref.png}};
	\spy [yellow] on (rel cs:x=40,y=10.0,name=img7) in node [below] at ($(img7.south)+(0,0.2)$);

	\node[inner sep=0,right=1pt of img2] (img8)
	{\includegraphics[height=1.75cm]{images/failure_cases/0ef68/ours32.png}};
	\spy [yellow] on (rel cs:x=40,y=10.0,name=img8) in node [below] at ($(img8.south)+(0,0.2)$);
	
	\node[inner sep=0,right=1pt of img3] (img9)
	{\includegraphics[height=1.75cm]{images/failure_cases/0ef68/3d_photo.png}};
	\spy [yellow] on (rel cs:x=40,y=10.0,name=img9) in node [below] at ($(img9.south)+(0,0.2)$);

	\node[inner sep=0,right=1pt of img4] (img10)
	{\includegraphics[height=1.75cm]{images/failure_cases/0ef68/ostereo_mag.png}};
	\spy [yellow] on (rel cs:x=40,y=10.0,name=img10) in node [below] at ($(img10.south)+(0,0.2)$);

	\end{tikzpicture}%
	\caption{Limitations of FaDIV-Syn: Inpainting at borders.}
	\label{fig:inpainting}\vspace{-2ex}
\end{figure*}

\begin{figure*}
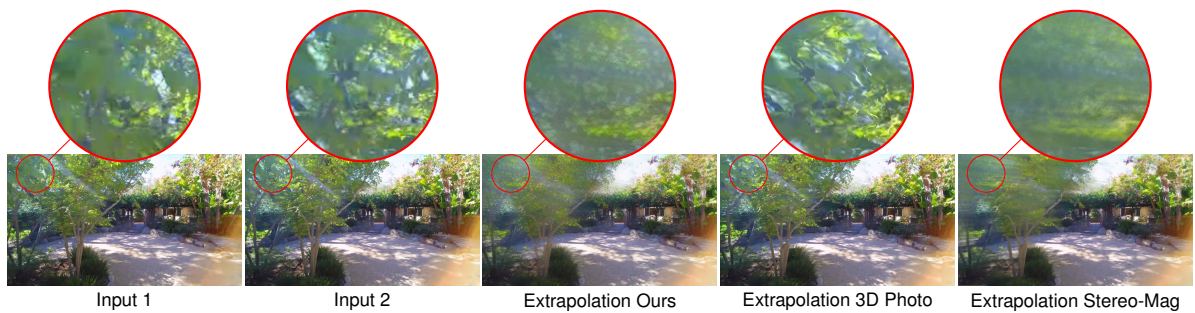
\sf \scriptsize \centering
	\begin{tikzpicture}[
	spy using outlines={circle, magnification=4, size=2cm, connect spies}
	]
	
	\node[inner sep=0] (img1)
	{\includegraphics[height=1.75cm]{images/failure_cases/1b29/source.png}};
	\spy [red] on (rel cs:x=12,y=85,name=img1) in node [above] at ($(img1.north)+(0,-0.1)$);
	
	\node[below] at (img1.south) {Input 1};
	
	\node[inner sep=0,right=1pt of img1] (img2)
	{\includegraphics[height=1.75cm]{images/failure_cases/1b29/ref.png}};
	\spy [red] on (rel cs:x=12,y=85,name=img2) in node [above] at ($(img2.north)+(0,-0.1)$);
	
	\node[below] at (img2.south) {Input 2};
	
	\node[inner sep=0,right=1pt of img2] (img3)
	{\includegraphics[height=1.75cm]{images/failure_cases/1b29/ours32.png}};
	\spy [red] on (rel cs:x=12,y=85,name=img3) in node [above] at ($(img3.north)+(0,-0.1)$);
	
	\node[below] at (img3.south) {Extrapolation Ours};
	
	\node[inner sep=0,right=1pt of img3] (img4)
	{\includegraphics[height=1.75cm]{images/failure_cases/1b29/3d_photo.png}};
	\spy [red] on (rel cs:x=12,y=85,name=img4) in node [above] at ($(img4.north)+(0,-0.1)$);
	
	\node[below] at (img4.south) {Extrapolation 3D Photo};
	
	\node[inner sep=0,right=1pt of img4] (img5)
	{\includegraphics[height=1.75cm]{images/failure_cases/1b29/ostereo_mag.png}};
	\spy [red] on (rel cs:x=12,y=85,name=img5) in node [above] at ($(img5.north)+(0,-0.1)$);
	
	\node[below] at (img5.south) {Extrapolation Stereo-Mag};
	\end{tikzpicture}

	\caption{Limitations of FaDIV-Syn: Insufficient pose alignment.}
	\label{fig:pose}\vspace{-2ex}
\end{figure*}

\begin{figure*}
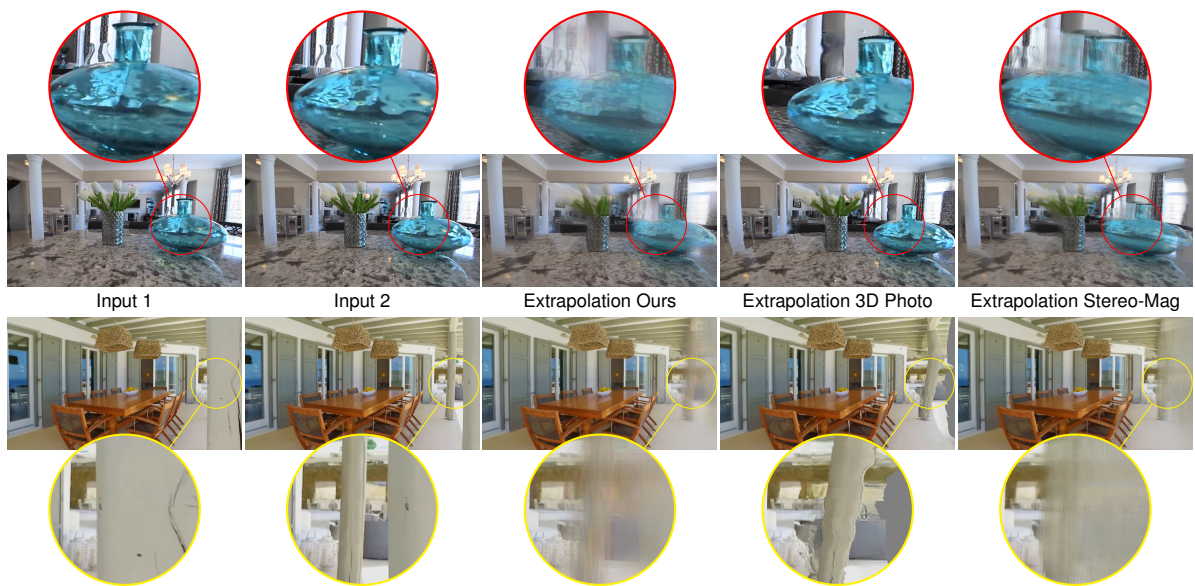
\sf \scriptsize \centering
	\begin{tikzpicture}[
	spy using outlines={circle, magnification=2.5, size=2cm, connect spies}
	]
	
	\node[inner sep=0] (img1)
	{\includegraphics[height=1.75cm]{images/failure_cases/12a/source.png}};
	\spy [red] on (rel cs:x=74,y=47,name=img1) in node [above] at ($(img1.north)+(0,-0.1)$);
	
	\node[below] at (img1.south) {Input 1};
	
	\node[inner sep=0,right=1pt of img1] (img2)
	{\includegraphics[height=1.75cm]{images/failure_cases/12a/ref.png}};
	\spy [red] on (rel cs:x=74,y=47,name=img2) in node [above] at ($(img2.north)+(0,-0.1)$);
	
	\node[below] at (img2.south) {Input 2};
	
	\node[inner sep=0,right=1pt of img2] (img3)
	{\includegraphics[height=1.75cm]{images/failure_cases/12a/ours32.png}};
	\spy [red] on (rel cs:x=74,y=47,name=img3) in node [above] at ($(img3.north)+(0,-0.1)$);
	
	\node[below] at (img3.south) {Extrapolation Ours};
	
	\node[inner sep=0,right=1pt of img3] (img4)
	{\includegraphics[height=1.75cm]{images/failure_cases/12a/3d_photo.png}};
	\spy [red] on (rel cs:x=74,y=47,name=img4) in node [above] at ($(img4.north)+(0,-0.1)$);
	
	\node[below] at (img4.south) {Extrapolation 3D Photo};
	
	\node[inner sep=0,right=1pt of img4] (img5)
	{\includegraphics[height=1.75cm]{images/failure_cases/12a/stereo_mag.png}};
	\spy [red] on (rel cs:x=74,y=47,name=img5) in node [above] at ($(img5.north)+(0,-0.1)$);
	
	\node[below] at (img5.south) {Extrapolation Stereo-Mag};
	\end{tikzpicture}
	
	\begin{tikzpicture}[
	spy using outlines={circle, magnification=3.0, size=2cm, connect spies}
	]
	
	\node[inner sep=0] (img6)
	{\includegraphics[height=1.75cm]{images/failure_cases/7b700/source.png}};
	\spy [yellow] on (rel cs:x=89,y=50.0,name=img1) in node [below] at ($(img6.south)+(0,0.2)$);

	\node[inner sep=0,right=1pt of img1] (img7)
	{\includegraphics[height=1.75cm]{images/failure_cases/7b700/ref.png}};
	\spy [yellow] on (rel cs:x=89,y=50.0,name=img7) in node [below] at ($(img7.south)+(0,0.2)$);

	\node[inner sep=0,right=1pt of img2] (img8)
	{\includegraphics[height=1.75cm]{images/failure_cases/7b700/ours32.png}};
	\spy [yellow] on (rel cs:x=89,y=50.0,name=img8) in node [below] at ($(img8.south)+(0,0.2)$);
	
	\node[inner sep=0,right=1pt of img3] (img9)
	{\includegraphics[height=1.75cm]{images/failure_cases/7b700/3d_photo.png}};
	\spy [yellow] on (rel cs:x=89,y=50.0,name=img9) in node [below] at ($(img9.south)+(0,0.2)$);
	
	\node[inner sep=0,right=1pt of img4] (img10)
	{\includegraphics[height=1.75cm]{images/failure_cases/7b700/ostereo_mag.png}};
	\spy [yellow] on (rel cs:x=89,y=50.0,name=img10) in node [below] at ($(img10.south)+(0,0.2)$);

	\end{tikzpicture}
	\caption{Limitations of FaDIV-Syn: General failure cases.}
	\label{fig:fail}\vspace{-2ex}
\end{figure*}

\paragraph{Biased training data}
Our method is trained in such a way that target poses are always on the camera trajectory of the
RealEstate10k dataset, where ground truth is available. However, this induces a bias, which may
result in less performance for target views outside of the smooth camera trajectories.
It could be possible to improve generalization to such poses using semi-supervised techniques~\cite{hani2020continuous}.

\paragraph{Slow convergence}
We note that the experiments in the main paper were achieved with a limited number of training iterations, i.e.,
we did not train until there was no improvement anymore.
\cref{tab:more_train} compares the accuracy of three equivalent networks which were trained for an ascending number of iterations on the test set provided by \cite{3D_photo}. \cref{tab:more_train} and \cref{fig:training_curve_corr19} illustrate that more training still improves the testing/validation accuracy and the optimal performance is not yet achieved.
It would be desirable to reach the optimal performance in significantly less training iterations
to achieve better results without changing the method itself.

\vspace{-0.1ex}
\section{Additional Qualitative Results}
\vspace{-0.1ex}
In addition to the exemplary results already shown, we present more qualitative examples here.
\Cref{fig:big_full,fig:big_full_2} show extrapolation examples for our full network variants (Ours-32, Ours-19) that use Soft-Masking,
in comparison to ground truth, 3D~Photo~\citep{3D_photo}, and Stereo-Mag~\citep{stereo_mag}.
In \cref{fig:full_generalized} we illustrate that our method ablation Ours-17-NoSM without Soft-Masking achieves also reasonable results.
To further demonstrate the generalization capability of our method to higher resolutions, we show interpolation sequences in \cref{fig:interpolation,fig:interpolation2} with models that are trained in 288p but inferred in 576p (twice the training resolution).

\tikzstyle{l}=[minimum width=3.4cm,inner sep=1pt,fill=yellow!50,chamfered rectangle,font=\sffamily\footnotesize]
\tikzstyle{ll}=[l,chamfered rectangle corners={south east, south west}]
\tikzstyle{ul}=[l,chamfered rectangle corners={north east, north west}]

\tikzstyle{input}=[fill=yellow!50,minimum width=3cm]
\tikzstyle{gt}=[fill=green!50]
\tikzstyle{p3d}=[fill=red!50]
\tikzstyle{full generalized}=[fill=cyan!50]
\tikzstyle{full}=[fill=blue!50]
\tikzstyle{big}=[fill=magenta!50]

\begin{figure*} \centering
 \begin{tikzpicture}[every node/.style={inner sep=0pt}, font=\sffamily,
    spy using outlines={magnification=3, size=1.5cm, connect spies,
      every spy on node/.append style={thick},
      spy connection path={\draw[thick] (tikzspyonnode) -- (tikzspyinnode);}
    }]
  \node (input1) {\includegraphics[height=1.7cm]{images/vgl/1a55c677aabd1a44/source.png}};
  \node[below=3pt of input1,draw=green,inner sep=1pt,ultra thick] (input2) {\includegraphics[height=1.7cm]{images/vgl/1a55c677aabd1a44/ref.png}};

  \node[anchor=south west] (out1) at ($(input2.south east)+(1pt,0)$) {\includegraphics[height=2cm]{images/vgl/1a55c677aabd1a44/ours32.png}};
  \node[right=1pt of out1] (out2) {\includegraphics[height=2cm]{images/vgl/1a55c677aabd1a44/gt.png}};
  \node[right=1pt of out2] (out3) {\includegraphics[height=2cm]{images/vgl/1a55c677aabd1a44/3d_photo.png}};
  \node[right=1pt of out3] (out4) {\includegraphics[height=2cm]{images/vgl/1a55c677aabd1a44/ostereo_mag.png}};

  \spy [cyan,magnification=2.8] on (rel cs:x=92.5,y=31,name=out1) in node [anchor=south west] at ($(out1.north west)+(0.1,0.1)$);
  \spy [cyan,magnification=2.8] on (rel cs:x=92.5,y=31,name=out2) in node [anchor=south west] at ($(out2.north west)+(0.1,0.1)$);
  \spy [cyan,magnification=2.8] on (rel cs:x=92.5,y=31,name=out3) in node [anchor=south west] at ($(out3.north west)+(0.1,0.1)$);
  \spy [cyan,magnification=2.8] on (rel cs:x=92.5,y=31,name=out4) in node [anchor=south west] at ($(out4.north west)+(0.1,0.1)$);

  \spy [magenta,magnification=2.8] on (rel cs:x=92.5,y=71,name=out1) in node [anchor=south east] at ($(out1.north east)+(-0.1,0.1)$);
  \spy [magenta,magnification=2.8] on (rel cs:x=92.5,y=71,name=out2) in node [anchor=south east] at ($(out2.north east)+(-0.1,0.1)$);
  \spy [magenta,magnification=2.8] on (rel cs:x=92.5,y=71,name=out3) in node [anchor=south east] at ($(out3.north east)+(-0.1,0.1)$);
  \spy [magenta,magnification=2.8] on (rel cs:x=92.5,y=71,name=out4) in node [anchor=south east] at ($(out4.north east)+(-0.1,0.1)$);

  \coordinate (labelpos) at ($(input1.north)+(0,0.3)$);

  \node[ul,input] at (labelpos-|input2) {Input};
  \node[ul,big] at (labelpos-|out1) {Ours-32};
  \node[ul,full] at (labelpos-|out2) {Ground Truth};
  \node[ul,gt] at (labelpos-|out3) {3D Photo~\citep{3D_photo}};
  \node[ul,p3d] at (labelpos-|out4) {Stereo-Mag~\citep{stereo_mag}};
 \end{tikzpicture}

 \vspace{7pt}
 
  \begin{tikzpicture}[every node/.style={inner sep=0pt}, font=\sffamily,
 spy using outlines={magnification=3, size=1.5cm, connect spies,
 	every spy on node/.append style={thick},
 	spy connection path={\draw[thick] (tikzspyonnode) -- (tikzspyinnode);}
 }]
 \node (input1) {\includegraphics[height=1.7cm]{images/vgl/4fcab7264432f232/source.png}};
 \node[below=3pt of input1,draw=green,inner sep=1pt,ultra thick] (input2) {\includegraphics[height=1.7cm]{images/vgl/4fcab7264432f232/ref.png}};
 
 \node[anchor=south west] (out1) at ($(input2.south east)+(1pt,0)$) {\includegraphics[height=2cm]{images/vgl/4fcab7264432f232/ours32.png}};
 \node[right=1pt of out1] (out2) {\includegraphics[height=2cm]{images/vgl/4fcab7264432f232/gt.png}};
 \node[right=1pt of out2] (out3) {\includegraphics[height=2cm]{images/vgl/4fcab7264432f232/3d_photo.png}};
 \node[right=1pt of out3] (out4) {\includegraphics[height=2cm]{images/vgl/4fcab7264432f232/ostereo_mag.png}};
 
 \spy [cyan,magnification=2.5] on (rel cs:x=40,y=15.8,name=out1) in node [anchor=south west] at ($(out1.north west)+(0.1,0.1)$);
 \spy [cyan,magnification=2.5] on (rel cs:x=40,y=15.8,name=out2) in node [anchor=south west] at ($(out2.north west)+(0.1,0.1)$);
 \spy [cyan,magnification=2.5] on (rel cs:x=40,y=15.8,name=out3) in node [anchor=south west] at ($(out3.north west)+(0.1,0.1)$);
 \spy [cyan,magnification=2.5] on (rel cs:x=40,y=15.8,name=out4) in node [anchor=south west] at ($(out4.north west)+(0.1,0.1)$);
 
 \spy [magenta,magnification=3] on (rel cs:x=71,y=60,name=out1) in node [anchor=south east] at ($(out1.north east)+(-0.1,0.1)$);
 \spy [magenta,magnification=3] on (rel cs:x=71,y=60,name=out2) in node [anchor=south east] at ($(out2.north east)+(-0.1,0.1)$);
 \spy [magenta,magnification=3] on (rel cs:x=71,y=60,name=out3) in node [anchor=south east] at ($(out3.north east)+(-0.1,0.1)$);
 \spy [magenta,magnification=3] on (rel cs:x=71,y=60,name=out4) in node [anchor=south east] at ($(out4.north east)+(-0.1,0.1)$);
 \end{tikzpicture}
 
  \vspace{7pt}
 
 \begin{tikzpicture}[every node/.style={inner sep=0pt}, font=\sffamily,
 spy using outlines={magnification=3, size=1.5cm, connect spies,
 	every spy on node/.append style={thick},
 	spy connection path={\draw[thick] (tikzspyonnode) -- (tikzspyinnode);}
 }]
 \node (input1) {\includegraphics[height=1.7cm]{images/vgl/9f57b4ea44c45ab9/source.png}};
 \node[below=3pt of input1,draw=green,inner sep=1pt,ultra thick] (input2) {\includegraphics[height=1.7cm]{images/vgl/9f57b4ea44c45ab9/ref.png}};
 
 \node[anchor=south west] (out1) at ($(input2.south east)+(1pt,0)$) {\includegraphics[height=2cm]{images/vgl/9f57b4ea44c45ab9/ours32.png}};
 \node[right=1pt of out1] (out2) {\includegraphics[height=2cm]{images/vgl/9f57b4ea44c45ab9/gt.png}};
 \node[right=1pt of out2] (out3) {\includegraphics[height=2cm]{images/vgl/9f57b4ea44c45ab9/3d_photo.png}};
 \node[right=1pt of out3] (out4) {\includegraphics[height=2cm]{images/vgl/9f57b4ea44c45ab9/ostereo_mag.png}};
 
 \spy [cyan,magnification=3.4] on (rel cs:x=43,y=42,name=out1) in node [anchor=south west] at ($(out1.north west)+(0.1,0.1)$);
 \spy [cyan,magnification=3.4] on (rel cs:x=43,y=42,name=out2) in node [anchor=south west] at ($(out2.north west)+(0.1,0.1)$);
 \spy [cyan,magnification=3.4] on (rel cs:x=43,y=42,name=out3) in node [anchor=south west] at ($(out3.north west)+(0.1,0.1)$);
 \spy [cyan,magnification=3.4] on (rel cs:x=43,y=42,name=out4) in node [anchor=south west] at ($(out4.north west)+(0.1,0.1)$);
 
 \spy [magenta,magnification=2.9] on (rel cs:x=62,y=37,name=out1) in node [anchor=south east] at ($(out1.north east)+(-0.1,0.1)$);
 \spy [magenta,magnification=2.9] on (rel cs:x=62,y=37,name=out2) in node [anchor=south east] at ($(out2.north east)+(-0.1,0.1)$);
 \spy [magenta,magnification=2.9] on (rel cs:x=62,y=37,name=out3) in node [anchor=south east] at ($(out3.north east)+(-0.1,0.1)$);
 \spy [magenta,magnification=2.9] on (rel cs:x=62,y=37,name=out4) in node [anchor=south east] at ($(out4.north east)+(-0.1,0.1)$);
 \end{tikzpicture}
 
 \vspace{7pt}
 
 \begin{tikzpicture}[every node/.style={inner sep=0pt}, font=\sffamily,
    spy using outlines={magnification=3, size=1.5cm, connect spies,
      every spy on node/.append style={thick},
      spy connection path={\draw[thick] (tikzspyonnode) -- (tikzspyinnode);}
    }]
  \node (input1) {\includegraphics[height=1.7cm]{images/vgl/5349781d9cdda05c/source.png}};
  \node[below=3pt of input1,draw=green,inner sep=1pt,ultra thick] (input2) {\includegraphics[height=1.7cm]{images/vgl/5349781d9cdda05c/ref.png}};

  \node[anchor=south west] (out1) at ($(input2.south east)+(1pt,0)$) {\includegraphics[height=2cm]{images/vgl/5349781d9cdda05c/ours32.png}};
  \node[right=1pt of out1] (out2) {\includegraphics[height=2cm]{images/vgl/5349781d9cdda05c/gt.png}};
  \node[right=1pt of out2] (out3) {\includegraphics[height=2cm]{images/vgl/5349781d9cdda05c/3d_photo.png}};
  \node[right=1pt of out3] (out4) {\includegraphics[height=2cm]{images/vgl/5349781d9cdda05c/ostereo_mag.png}};

  \spy [cyan,magnification=1.9] on (rel cs:x=32,y=30,name=out1) in node [anchor=south west] at ($(out1.north west)+(0.1,0.1)$);
  \spy [cyan,magnification=1.9] on (rel cs:x=32,y=30,name=out2) in node [anchor=south west] at ($(out2.north west)+(0.1,0.1)$);
  \spy [cyan,magnification=1.9] on (rel cs:x=32,y=30,name=out3) in node [anchor=south west] at ($(out3.north west)+(0.1,0.1)$);
  \spy [cyan,magnification=1.9] on (rel cs:x=32,y=30,name=out4) in node [anchor=south west] at ($(out4.north west)+(0.1,0.1)$);

  \spy [magenta,magnification=1.7] on (rel cs:x=65,y=40,name=out1) in node [anchor=south east] at ($(out1.north east)+(-0.1,0.1)$);
  \spy [magenta,magnification=1.7] on (rel cs:x=65,y=40,name=out2) in node [anchor=south east] at ($(out2.north east)+(-0.1,0.1)$);
  \spy [magenta,magnification=1.7] on (rel cs:x=65,y=40,name=out3) in node [anchor=south east] at ($(out3.north east)+(-0.1,0.1)$);
  \spy [magenta,magnification=1.7] on (rel cs:x=65,y=40,name=out4) in node [anchor=south east] at ($(out4.north east)+(-0.1,0.1)$);
 \end{tikzpicture}

 \vspace{7pt}

 \begin{tikzpicture}[every node/.style={inner sep=0pt}, font=\sffamily,
    spy using outlines={magnification=3, size=1.5cm, connect spies,
      every spy on node/.append style={thick},
      spy connection path={\draw[thick] (tikzspyonnode) -- (tikzspyinnode);}
    }]
  \node (input1) {\includegraphics[height=1.7cm]{images/vgl/e84891165f0bf125/source.png}};
  \node[below=3pt of input1,draw=green,inner sep=1pt,ultra thick] (input2) {\includegraphics[height=1.7cm]{images/vgl/e84891165f0bf125/ref.png}};

  \node[anchor=south west] (out1) at ($(input2.south east)+(1pt,0)$) {\includegraphics[height=2cm]{images/vgl/e84891165f0bf125/ours32.png}};
  \node[right=1pt of out1] (out2) {\includegraphics[height=2cm]{images/vgl/e84891165f0bf125/gt.png}};
  \node[right=1pt of out2] (out3) {\includegraphics[height=2cm]{images/vgl/e84891165f0bf125/3d_photo.png}};
  \node[right=1pt of out3] (out4) {\includegraphics[height=2cm]{images/vgl/e84891165f0bf125/ostereo_mag.png}};

  \spy [cyan,magnification=2.6] on (rel cs:x=40,y=15,name=out1) in node [anchor=south west] at ($(out1.north west)+(0.1,0.1)$);
  \spy [cyan,magnification=2.6] on (rel cs:x=40,y=15,name=out2) in node [anchor=south west] at ($(out2.north west)+(0.1,0.1)$);
  \spy [cyan,magnification=2.6] on (rel cs:x=40,y=15,name=out3) in node [anchor=south west] at ($(out3.north west)+(0.1,0.1)$);
  \spy [cyan,magnification=2.6] on (rel cs:x=40,y=15,name=out4) in node [anchor=south west] at ($(out4.north west)+(0.1,0.1)$);

  \spy [magenta,magnification=2.2] on (rel cs:x=45,y=60,name=out1) in node [anchor=south east] at ($(out1.north east)+(-0.1,0.1)$);
  \spy [magenta,magnification=2.2] on (rel cs:x=45,y=60,name=out2) in node [anchor=south east] at ($(out2.north east)+(-0.1,0.1)$);
  \spy [magenta,magnification=2.2] on (rel cs:x=45,y=60,name=out3) in node [anchor=south east] at ($(out3.north east)+(-0.1,0.1)$);
  \spy [magenta,magnification=2.2] on (rel cs:x=45,y=60,name=out4) in node [anchor=south east] at ($(out4.north east)+(-0.1,0.1)$);
  \coordinate (labelpos) at ($(input2.south)-(0,0.3)$);
  
  \node[ll,input] at (labelpos-|input2) {Input};
  \node[ll,big] at (labelpos-|out1) {Ours-32};
  \node[ll,full] at (labelpos-|out2) {Ground Truth};
  \node[ll,gt] at (labelpos-|out3) {3D Photo~\citep{3D_photo}};
  \node[ll,p3d] at (labelpos-|out4) {Stereo-Mag~\cite{stereo_mag}};
 \end{tikzpicture}

 \caption{Extrapolation comparison of our method with 32 planes (Ours-32) against ground truth, 3D Photo~\citep{3D_photo}, and Stereo-Mag~\cite{stereo_mag}.
  The input frame closer to the target frame is marked in green for easier comparison.}
  \label{fig:big_full}
\end{figure*}

\begin{figure*} \centering
	\begin{tikzpicture}[every node/.style={inner sep=0pt}, font=\sffamily,
	spy using outlines={magnification=3, size=1.5cm, connect spies,
		every spy on node/.append style={thick},
		spy connection path={\draw[thick] (tikzspyonnode) -- (tikzspyinnode);}
	}]
	\node (input1) {\includegraphics[height=1.7cm]{images/vgl/2e06abf6286040e2/source.png}};
	\node[below=3pt of input1,draw=green,inner sep=1pt,ultra thick] (input2) {\includegraphics[height=1.7cm]{images/vgl/2e06abf6286040e2/ref.png}};
	
	\node[anchor=south west] (out1) at ($(input2.south east)+(1pt,0)$) {\includegraphics[height=2cm]{images/vgl/2e06abf6286040e2/ours19.png}};
	\node[right=1pt of out1] (out2) {\includegraphics[height=2cm]{images/vgl/2e06abf6286040e2/gt.png}};
	\node[right=1pt of out2] (out3) {\includegraphics[height=2cm]{images/vgl/2e06abf6286040e2/3d_photo.png}};
	\node[right=1pt of out3] (out4) {\includegraphics[height=2cm]{images/vgl/2e06abf6286040e2/ostereo_mag.png}};
	
	\spy [cyan,magnification=4.5] on (rel cs:x=12,y=91,name=out1) in node [anchor=south west] at ($(out1.north west)+(0.1,0.1)$);
	\spy [cyan,magnification=4.5] on (rel cs:x=12,y=91,name=out2) in node [anchor=south west] at ($(out2.north west)+(0.1,0.1)$);
	\spy [cyan,magnification=4.5] on (rel cs:x=12,y=91,name=out3) in node [anchor=south west] at ($(out3.north west)+(0.1,0.1)$);
	\spy [cyan,magnification=4.5] on (rel cs:x=12,y=91,name=out4) in node [anchor=south west] at ($(out4.north west)+(0.1,0.1)$);
	
	\spy [magenta,magnification=6.6] on (rel cs:x=85.5,y=64,name=out1) in node [anchor=south east] at ($(out1.north east)+(-0.1,0.1)$);
	\spy [magenta,magnification=6.6] on (rel cs:x=85.5,y=64,name=out2) in node [anchor=south east] at ($(out2.north east)+(-0.1,0.1)$);
	\spy [magenta,magnification=6.6] on (rel cs:x=85.5,y=64,name=out3) in node [anchor=south east] at ($(out3.north east)+(-0.1,0.1)$);
	\spy [magenta,magnification=6.6] on (rel cs:x=85.5,y=64,name=out4) in node [anchor=south east] at ($(out4.north east)+(-0.1,0.1)$);
	
	\coordinate (labelpos) at ($(input1.north)+(0,0.3)$);
	
	\node[ul,input] at (labelpos-|input2) {Input};
  \node[ul,big] at (labelpos-|out1) {Ours-19};
  \node[ul,full] at (labelpos-|out2) {Ground Truth};
  \node[ul,gt] at (labelpos-|out3) {3D Photo~\cite{3D_photo}};
  \node[ul,p3d] at (labelpos-|out4) {Stereo-Mag~\cite{stereo_mag}};
	\end{tikzpicture}
	
	\vspace{7pt}

	\begin{tikzpicture}[every node/.style={inner sep=0pt}, font=\sffamily,
	spy using outlines={magnification=3, size=1.5cm, connect spies,
		every spy on node/.append style={thick},
		spy connection path={\draw[thick] (tikzspyonnode) -- (tikzspyinnode);}
	}]
	\node (input1) {\includegraphics[height=1.7cm]{images/vgl/9b3d6bbe5ee18684/source.png}};
	\node[below=3pt of input1,draw=green,inner sep=1pt,ultra thick] (input2) {\includegraphics[height=1.7cm]{images/vgl/9b3d6bbe5ee18684/ref.png}};
	
	\node[anchor=south west] (out1) at ($(input2.south east)+(1pt,0)$) {\includegraphics[height=2cm]{images/vgl/9b3d6bbe5ee18684/ours19.png}};
	\node[right=1pt of out1] (out2) {\includegraphics[height=2cm]{images/vgl/9b3d6bbe5ee18684/gt.png}};
	\node[right=1pt of out2] (out3) {\includegraphics[height=2cm]{images/vgl/9b3d6bbe5ee18684/3d_photo.png}};
	\node[right=1pt of out3] (out4) {\includegraphics[height=2cm]{images/vgl/9b3d6bbe5ee18684/ostereo_mag.png}};
	
	\spy [cyan,magnification=3.15] on (rel cs:x=30,y=65,name=out1) in node [anchor=south west] at ($(out1.north west)+(0.1,0.1)$);
	\spy [cyan,magnification=3.15] on (rel cs:x=30,y=65,name=out2) in node [anchor=south west] at ($(out2.north west)+(0.1,0.1)$);
	\spy [cyan,magnification=3.15] on (rel cs:x=30,y=65,name=out3) in node [anchor=south west] at ($(out3.north west)+(0.1,0.1)$);
	\spy [cyan,magnification=3.15] on (rel cs:x=30,y=65,name=out4) in node [anchor=south west] at ($(out4.north west)+(0.1,0.1)$);
	
	\spy [magenta,magnification=2.8] on (rel cs:x=28,y=17,name=out1) in node [anchor=south east] at ($(out1.north east)+(-0.1,0.1)$);
	\spy [magenta,magnification=2.8] on (rel cs:x=28,y=17,name=out2) in node [anchor=south east] at ($(out2.north east)+(-0.1,0.1)$);
	\spy [magenta,magnification=2.8] on (rel cs:x=28,y=17,name=out3) in node [anchor=south east] at ($(out3.north east)+(-0.1,0.1)$);
	\spy [magenta,magnification=2.8] on (rel cs:x=28,y=17,name=out4) in node [anchor=south east] at ($(out4.north east)+(-0.1,0.1)$);
	\end{tikzpicture}
	
	\vspace{7pt}
	
	\begin{tikzpicture}[every node/.style={inner sep=0pt}, font=\sffamily,
	spy using outlines={magnification=3, size=1.5cm, connect spies,
		every spy on node/.append style={thick},
		spy connection path={\draw[thick] (tikzspyonnode) -- (tikzspyinnode);}
	}]
	\node (input1) {\includegraphics[height=1.7cm]{images/vgl/1b14e662c17cb551/source.png}};
	\node[below=3pt of input1,draw=green,inner sep=1pt,ultra thick] (input2) {\includegraphics[height=1.7cm]{images/vgl/1b14e662c17cb551/ref.png}};
	
	\node[anchor=south west] (out1) at ($(input2.south east)+(1pt,0)$) {\includegraphics[height=2cm]{images/vgl/1b14e662c17cb551/ours19.png}};
	\node[right=1pt of out1] (out2) {\includegraphics[height=2cm]{images/vgl/1b14e662c17cb551/gt.png}};
	\node[right=1pt of out2] (out3) {\includegraphics[height=2cm]{images/vgl/1b14e662c17cb551/3d_photo.png}};
	\node[right=1pt of out3] (out4) {\includegraphics[height=2cm]{images/vgl/1b14e662c17cb551/ostereo_mag.png}};
	
	\spy [cyan,magnification=4.4] on (rel cs:x=38,y=53,name=out1) in node [anchor=south west] at ($(out1.north west)+(0.1,0.1)$);
	\spy [cyan,magnification=4.4] on (rel cs:x=38,y=53,name=out2) in node [anchor=south west] at ($(out2.north west)+(0.1,0.1)$);
	\spy [cyan,magnification=4.4] on (rel cs:x=38,y=53,name=out3) in node [anchor=south west] at ($(out3.north west)+(0.1,0.1)$);
	\spy [cyan,magnification=4.4] on (rel cs:x=38,y=53,name=out4) in node [anchor=south west] at ($(out4.north west)+(0.1,0.1)$);
	
	\spy [magenta,magnification=6] on (rel cs:x=73,y=35,name=out1) in node [anchor=south east] at ($(out1.north east)+(-0.1,0.1)$);
	\spy [magenta,magnification=6] on (rel cs:x=73,y=35,name=out2) in node [anchor=south east] at ($(out2.north east)+(-0.1,0.1)$);
	\spy [magenta,magnification=6] on (rel cs:x=73,y=35,name=out3) in node [anchor=south east] at ($(out3.north east)+(-0.1,0.1)$);
	\spy [magenta,magnification=6] on (rel cs:x=73,y=35,name=out4) in node [anchor=south east] at ($(out4.north east)+(-0.1,0.1)$);
	\end{tikzpicture}
	
	\vspace{7pt}
	
	\begin{tikzpicture}[every node/.style={inner sep=0pt}, font=\sffamily,
	spy using outlines={magnification=3, size=1.5cm, connect spies,
		every spy on node/.append style={thick},
		spy connection path={\draw[thick] (tikzspyonnode) -- (tikzspyinnode);}
	}]
	\node (input1) {\includegraphics[height=1.7cm]{images/vgl/8cdbbd2eca6ba5df/source.png}};
	\node[below=3pt of input1,draw=green,inner sep=1pt,ultra thick] (input2) {\includegraphics[height=1.7cm]{images/vgl/8cdbbd2eca6ba5df/ref.png}};
	
	\node[anchor=south west] (out1) at ($(input2.south east)+(1pt,0)$) {\includegraphics[height=2cm]{images/vgl/8cdbbd2eca6ba5df/ours19.png}};
	\node[right=1pt of out1] (out2) {\includegraphics[height=2cm]{images/vgl/8cdbbd2eca6ba5df/gt.png}};
	\node[right=1pt of out2] (out3) {\includegraphics[height=2cm]{images/vgl/8cdbbd2eca6ba5df/3d_photo.png}};
	\node[right=1pt of out3] (out4) {\includegraphics[height=2cm]{images/vgl/8cdbbd2eca6ba5df/ostereo_mag.png}};
	
	\spy [cyan,magnification=2.6] on (rel cs:x=28,y=65,name=out1) in node [anchor=south west] at ($(out1.north west)+(0.1,0.1)$);
	\spy [cyan,magnification=2.6] on (rel cs:x=28,y=65,name=out2) in node [anchor=south west] at ($(out2.north west)+(0.1,0.1)$);
	\spy [cyan,magnification=2.6] on (rel cs:x=28,y=65,name=out3) in node [anchor=south west] at ($(out3.north west)+(0.1,0.1)$);
	\spy [cyan,magnification=2.6] on (rel cs:x=28,y=65,name=out4) in node [anchor=south west] at ($(out4.north west)+(0.1,0.1)$);
	
	\spy [magenta,magnification=2.6] on (rel cs:x=90.5,y=63,name=out1) in node [anchor=south east] at ($(out1.north east)+(-0.1,0.1)$);
	\spy [magenta,magnification=2.6] on (rel cs:x=90.5,y=63,name=out2) in node [anchor=south east] at ($(out2.north east)+(-0.1,0.1)$);
	\spy [magenta,magnification=2.6] on (rel cs:x=90.5,y=63,name=out3) in node [anchor=south east] at ($(out3.north east)+(-0.1,0.1)$);
	\spy [magenta,magnification=2.6] on (rel cs:x=90.5,y=63,name=out4) in node [anchor=south east] at ($(out4.north east)+(-0.1,0.1)$);
	\end{tikzpicture}
	
	\vspace{7pt}
	
	\begin{tikzpicture}[every node/.style={inner sep=0pt}, font=\sffamily,
	spy using outlines={magnification=3, size=1.5cm, connect spies,
		every spy on node/.append style={thick},
		spy connection path={\draw[thick] (tikzspyonnode) -- (tikzspyinnode);}
	}]
	\node (input1) {\includegraphics[height=1.7cm]{images/vgl/987250887159290b/source.png}};
	\node[below=3pt of input1,draw=green,inner sep=1pt,ultra thick] (input2) {\includegraphics[height=1.7cm]{images/vgl/987250887159290b/ref.png}};
	
	\node[anchor=south west] (out1) at ($(input2.south east)+(1pt,0)$) {\includegraphics[height=2cm]{images/vgl/987250887159290b/ours19.png}};
	\node[right=1pt of out1] (out2) {\includegraphics[height=2cm]{images/vgl/987250887159290b/gt.png}};
	\node[right=1pt of out2] (out3) {\includegraphics[height=2cm]{images/vgl/987250887159290b/3d_photo.png}};
	\node[right=1pt of out3] (out4) {\includegraphics[height=2cm]{images/vgl/987250887159290b/ostereo_mag.png}};
	
	\spy [cyan,magnification=9.5] on (rel cs:x=46,y=64.5,name=out1) in node [anchor=south west] at ($(out1.north west)+(0.1,0.1)$);
	\spy [cyan,magnification=9.5] on (rel cs:x=46,y=64.5,name=out2) in node [anchor=south west] at ($(out2.north west)+(0.1,0.1)$);
	\spy [cyan,magnification=9.5] on (rel cs:x=46,y=64.5,name=out3) in node [anchor=south west] at ($(out3.north west)+(0.1,0.1)$);
	\spy [cyan,magnification=9.5] on (rel cs:x=46,y=64.5,name=out4) in node [anchor=south west] at ($(out4.north west)+(0.1,0.1)$);
	
	\spy [magenta,magnification=3] on (rel cs:x=79,y=45.7,name=out1) in node [anchor=south east] at ($(out1.north east)+(-0.1,0.1)$);
	\spy [magenta,magnification=3] on (rel cs:x=79,y=45.7,name=out2) in node [anchor=south east] at ($(out2.north east)+(-0.1,0.1)$);
	\spy [magenta,magnification=3] on (rel cs:x=79,y=45.7,name=out3) in node [anchor=south east] at ($(out3.north east)+(-0.1,0.1)$);
	\spy [magenta,magnification=3] on (rel cs:x=79,y=45.7,name=out4) in node [anchor=south east] at ($(out4.north east)+(-0.1,0.1)$);
	\coordinate (labelpos) at ($(input2.south)-(0,0.3)$);

\node[ll,input] at (labelpos-|input2) {Input};
\node[ll,big] at (labelpos-|out1) {Ours-19};
\node[ll,full] at (labelpos-|out2) {Ground Truth};
\node[ll,gt] at (labelpos-|out3) {3D Photo~\citep{3D_photo}};
\node[ll,p3d] at (labelpos-|out4) {Stereo-Mag~\cite{stereo_mag}};
	\end{tikzpicture}

	\caption{Extrapolation comparison of our method with 19 planes (Ours-19) against ground truth, 3D Photo~\citep{3D_photo}, and Stereo-Mag~\cite{stereo_mag}.
		The input frame closer to the target frame is marked in green for easier comparison.}
    \label{fig:big_full_2}
\end{figure*}

\begin{figure*}
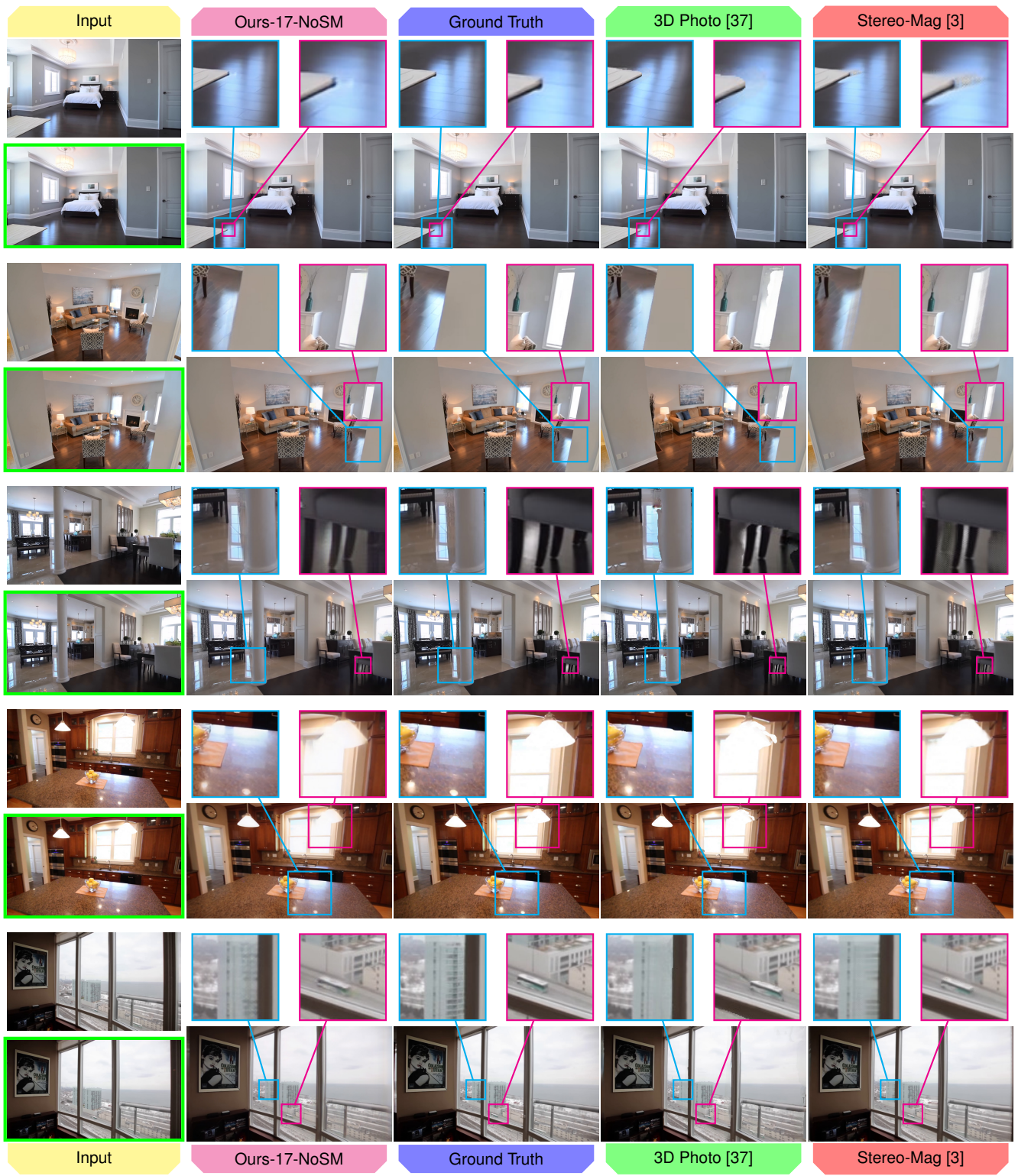
 \centering
	\begin{tikzpicture}[every node/.style={inner sep=0pt}, font=\sffamily\footnotesize,
	spy using outlines={magnification=3, size=1.5cm, connect spies,
		every spy on node/.append style={thick},
		spy connection path={\draw[thick] (tikzspyonnode) -- (tikzspyinnode);}
	}]
	\node (input1) {\includegraphics[height=1.7cm]{images/vgl/1aa94a69594df8a6/source.png}};
	\node[below=3pt of input1,draw=green,inner sep=1pt,ultra thick] (input2) {\includegraphics[height=1.7cm]{images/vgl/1aa94a69594df8a6/ref.png}};
	
	\node[anchor=south west] (out1) at ($(input2.south east)+(1pt,0)$) {\includegraphics[height=2cm]{images/vgl/1aa94a69594df8a6/finetuned.png}};
	\node[right=1pt of out1] (out2) {\includegraphics[height=2cm]{images/vgl/1aa94a69594df8a6/gt.png}};
	\node[right=1pt of out2] (out3) {\includegraphics[height=2cm]{images/vgl/1aa94a69594df8a6/3d_photo.png}};
	\node[right=1pt of out3] (out4) {\includegraphics[height=2cm]{images/vgl/1aa94a69594df8a6/ostereo_mag.png}};
	
	\spy [cyan,magnification=2.85] on (rel cs:x=21,y=13,name=out1) in node [anchor=south west] at ($(out1.north west)+(0.1,0.1)$);
	\spy [cyan,magnification=2.85] on (rel cs:x=21,y=13,name=out2) in node [anchor=south west] at ($(out2.north west)+(0.1,0.1)$);
	\spy [cyan,magnification=2.85] on (rel cs:x=21,y=13,name=out3) in node [anchor=south west] at ($(out3.north west)+(0.1,0.1)$);
	\spy [cyan,magnification=2.85] on (rel cs:x=21,y=13,name=out4) in node [anchor=south west] at ($(out4.north west)+(0.1,0.1)$);
	
	\spy [magenta,magnification=6.875] on (rel cs:x=20.5,y=16,name=out1) in node [anchor=south east] at ($(out1.north east)+(-0.1,0.1)$);
	\spy [magenta,magnification=6.875] on (rel cs:x=20.5,y=16,name=out2) in node [anchor=south east] at ($(out2.north east)+(-0.1,0.1)$);
	\spy [magenta,magnification=6.875] on (rel cs:x=20.5,y=16,name=out3) in node [anchor=south east] at ($(out3.north east)+(-0.1,0.1)$);
	\spy [magenta,magnification=6.875] on (rel cs:x=20.5,y=16,name=out4) in node [anchor=south east] at ($(out4.north east)+(-0.1,0.1)$);
	
	\coordinate (labelpos) at ($(input1.north)+(0,0.3)$);
	
	\node[ul,input] at (labelpos-|input2) {Input};
\node[ul,big] at (labelpos-|out1) {Ours-17-NoSM};
\node[ul,full] at (labelpos-|out2) {Ground Truth};
\node[ul,gt] at (labelpos-|out3) {3D Photo~\cite{3D_photo}};
\node[ul,p3d] at (labelpos-|out4) {Stereo-Mag~\cite{stereo_mag}};
	\end{tikzpicture}
	
	\vspace{7pt}
	\begin{tikzpicture}[every node/.style={inner sep=0pt}, font=\sffamily,
	spy using outlines={magnification=3, size=1.5cm, connect spies,
		every spy on node/.append style={thick},
		spy connection path={\draw[thick] (tikzspyonnode) -- (tikzspyinnode);}
	}]
	\node (input1) {\includegraphics[height=1.7cm]{images/vgl/0bcef9ed1c18f74d/source.png}};
	\node[below=3pt of input1,draw=green,inner sep=1pt,ultra thick] (input2) {\includegraphics[height=1.7cm]{images/vgl/0bcef9ed1c18f74d/ref.png}};
	
	\node[anchor=south west] (out1) at ($(input2.south east)+(1pt,0)$) {\includegraphics[height=2cm]{images/vgl/0bcef9ed1c18f74d/finetuned.png}};
	\node[right=1pt of out1] (out2) {\includegraphics[height=2cm]{images/vgl/0bcef9ed1c18f74d/gt.png}};
	\node[right=1pt of out2] (out3) {\includegraphics[height=2cm]{images/vgl/0bcef9ed1c18f74d/3d_photo.png}};
	\node[right=1pt of out3] (out4) {\includegraphics[height=2cm]{images/vgl/0bcef9ed1c18f74d/ostereo_mag.png}};
	
	\spy [cyan,magnification=2.5] on (rel cs:x=86,y=24,name=out1) in node [anchor=south west] at ($(out1.north west)+(0.1,0.1)$);
	\spy [cyan,magnification=2.5] on (rel cs:x=86,y=24,name=out2) in node [anchor=south west] at ($(out2.north west)+(0.1,0.1)$);
	\spy [cyan,magnification=2.5] on (rel cs:x=86,y=24,name=out3) in node [anchor=south west] at ($(out3.north west)+(0.1,0.1)$);
	\spy [cyan,magnification=2.5] on (rel cs:x=86,y=24,name=out4) in node [anchor=south west] at ($(out4.north west)+(0.1,0.1)$);
	
	\spy [magenta,magnification=2.3] on (rel cs:x=86,y=61,name=out1) in node [anchor=south east] at ($(out1.north east)+(-0.1,0.1)$);
	\spy [magenta,magnification=2.3] on (rel cs:x=86,y=61,name=out2) in node [anchor=south east] at ($(out2.north east)+(-0.1,0.1)$);
	\spy [magenta,magnification=2.3] on (rel cs:x=86,y=61,name=out3) in node [anchor=south east] at ($(out3.north east)+(-0.1,0.1)$);
	\spy [magenta,magnification=2.3] on (rel cs:x=86,y=61,name=out4) in node [anchor=south east] at ($(out4.north east)+(-0.1,0.1)$);
	\end{tikzpicture}
	
	\vspace{7pt}
	
	\begin{tikzpicture}[every node/.style={inner sep=0pt}, font=\sffamily,
	spy using outlines={magnification=3, size=1.5cm, connect spies,
		every spy on node/.append style={thick},
		spy connection path={\draw[thick] (tikzspyonnode) -- (tikzspyinnode);}
	}]
	\node (input1) {\includegraphics[height=1.7cm]{images/vgl/0278b3d8abd9654d/source.png}};
	\node[below=3pt of input1,draw=green,inner sep=1pt,ultra thick] (input2) {\includegraphics[height=1.7cm]{images/vgl/0278b3d8abd9654d/ref.png}};
	
	\node[anchor=south west] (out1) at ($(input2.south east)+(1pt,0)$) {\includegraphics[height=2cm]{images/vgl/0278b3d8abd9654d/finetuned.png}};
	\node[right=1pt of out1] (out2) {\includegraphics[height=2cm]{images/vgl/0278b3d8abd9654d/gt.png}};
	\node[right=1pt of out2] (out3) {\includegraphics[height=2cm]{images/vgl/0278b3d8abd9654d/3d_photo.png}};
	\node[right=1pt of out3] (out4) {\includegraphics[height=2cm]{images/vgl/0278b3d8abd9654d/ostereo_mag.png}};
	
	\spy [cyan,magnification=2.5] on (rel cs:x=30,y=26,name=out1) in node [anchor=south west] at ($(out1.north west)+(0.1,0.1)$);
	\spy [cyan,magnification=2.5] on (rel cs:x=30,y=26,name=out2) in node [anchor=south west] at ($(out2.north west)+(0.1,0.1)$);
	\spy [cyan,magnification=2.5] on (rel cs:x=30,y=26,name=out3) in node [anchor=south west] at ($(out3.north west)+(0.1,0.1)$);
	\spy [cyan,magnification=2.5] on (rel cs:x=30,y=26,name=out4) in node [anchor=south west] at ($(out4.north west)+(0.1,0.1)$);
	
	\spy [magenta,magnification=5.5] on (rel cs:x=86,y=26,name=out1) in node [anchor=south east] at ($(out1.north east)+(-0.1,0.1)$);
	\spy [magenta,magnification=5.5] on (rel cs:x=86,y=26,name=out2) in node [anchor=south east] at ($(out2.north east)+(-0.1,0.1)$);
	\spy [magenta,magnification=5.5] on (rel cs:x=86,y=26,name=out3) in node [anchor=south east] at ($(out3.north east)+(-0.1,0.1)$);
	\spy [magenta,magnification=5.5] on (rel cs:x=86,y=26,name=out4) in node [anchor=south east] at ($(out4.north east)+(-0.1,0.1)$);
	\end{tikzpicture}
	
	\vspace{7pt}
 \begin{tikzpicture}[every node/.style={inner sep=0pt}, font=\sffamily,
spy using outlines={magnification=3, size=1.5cm, connect spies,
	every spy on node/.append style={thick},
	spy connection path={\draw[thick] (tikzspyonnode) -- (tikzspyinnode);}
}]
\node (input1) {\includegraphics[height=1.7cm]{images/vgl/01a628e2c509b823/source.png}};
\node[below=3pt of input1,draw=green,inner sep=1pt,ultra thick] (input2) {\includegraphics[height=1.7cm]{images/vgl/01a628e2c509b823/ref.png}};

\node[anchor=south west] (out1) at ($(input2.south east)+(1pt,0)$) {\includegraphics[height=2cm]{images/vgl/01a628e2c509b823/finetuned.png}};
\node[right=1pt of out1] (out2) {\includegraphics[height=2cm]{images/vgl/01a628e2c509b823/gt.png}};
\node[right=1pt of out2] (out3) {\includegraphics[height=2cm]{images/vgl/01a628e2c509b823/3d_photo.png}};
\node[right=1pt of out3] (out4) {\includegraphics[height=2cm]{images/vgl/01a628e2c509b823/ostereo_mag.png}};

\spy [cyan,magnification=2.0] on (rel cs:x=60,y=22,name=out1) in node [anchor=south west] at ($(out1.north west)+(0.1,0.1)$);
\spy [cyan,magnification=2.0] on (rel cs:x=60,y=22,name=out2) in node [anchor=south west] at ($(out2.north west)+(0.1,0.1)$);
\spy [cyan,magnification=2.0] on (rel cs:x=60,y=22,name=out3) in node [anchor=south west] at ($(out3.north west)+(0.1,0.1)$);
\spy [cyan,magnification=2.0] on (rel cs:x=60,y=22,name=out4) in node [anchor=south west] at ($(out4.north west)+(0.1,0.1)$);

\spy [magenta,magnification=2] on (rel cs:x=70,y=80,name=out1) in node [anchor=south east] at ($(out1.north east)+(-0.1,0.1)$);
\spy [magenta,magnification=2] on (rel cs:x=70,y=80,name=out2) in node [anchor=south east] at ($(out2.north east)+(-0.1,0.1)$);
\spy [magenta,magnification=2] on (rel cs:x=70,y=80,name=out3) in node [anchor=south east] at ($(out3.north east)+(-0.1,0.1)$);
\spy [magenta,magnification=2] on (rel cs:x=70,y=80,name=out4) in node [anchor=south east] at ($(out4.north east)+(-0.1,0.1)$);
	\end{tikzpicture}
	
	\vspace{7pt}
	
	\begin{tikzpicture}[every node/.style={inner sep=0pt}, font=\sffamily,
	spy using outlines={magnification=3, size=1.5cm, connect spies,
		every spy on node/.append style={thick},
		spy connection path={\draw[thick] (tikzspyonnode) -- (tikzspyinnode);}
	}]
	\node (input1) {\includegraphics[height=1.7cm]{images/vgl/750ddf09bd6d1eab/source.png}};
	\node[below=3pt of input1,draw=green,inner sep=1pt,ultra thick] (input2) {\includegraphics[height=1.7cm]{images/vgl/750ddf09bd6d1eab/ref.png}};
	
	\node[anchor=south west] (out1) at ($(input2.south east)+(1pt,0)$) {\includegraphics[height=2cm]{images/vgl/750ddf09bd6d1eab/finetuned.png}};
	\node[right=1pt of out1] (out2) {\includegraphics[height=2cm]{images/vgl/750ddf09bd6d1eab/gt.png}};
	\node[right=1pt of out2] (out3) {\includegraphics[height=2cm]{images/vgl/750ddf09bd6d1eab/3d_photo.png}};
	\node[right=1pt of out3] (out4) {\includegraphics[height=2cm]{images/vgl/750ddf09bd6d1eab/ostereo_mag.png}};
	
	\spy [cyan,magnification=4.5] on (rel cs:x=40,y=45,name=out1) in node [anchor=south west] at ($(out1.north west)+(0.1,0.1)$);
	\spy [cyan,magnification=4.5] on (rel cs:x=40,y=45,name=out2) in node [anchor=south west] at ($(out2.north west)+(0.1,0.1)$);
	\spy [cyan,magnification=4.5] on (rel cs:x=40,y=45,name=out3) in node [anchor=south west] at ($(out3.north west)+(0.1,0.1)$);
	\spy [cyan,magnification=4.5] on (rel cs:x=40,y=45,name=out4) in node [anchor=south west] at ($(out4.north west)+(0.1,0.1)$);
	
	\spy [magenta,magnification=4.5] on (rel cs:x=51,y=24,name=out1) in node [anchor=south east] at ($(out1.north east)+(-0.1,0.1)$);
	\spy [magenta,magnification=4.5] on (rel cs:x=51,y=24,name=out2) in node [anchor=south east] at ($(out2.north east)+(-0.1,0.1)$);
	\spy [magenta,magnification=4.5] on (rel cs:x=51,y=24,name=out3) in node [anchor=south east] at ($(out3.north east)+(-0.1,0.1)$);
	\spy [magenta,magnification=4.5] on (rel cs:x=51,y=24,name=out4) in node [anchor=south east] at ($(out4.north east)+(-0.1,0.1)$);
	
	\coordinate (labelpos) at ($(input2.south)-(0,0.3)$);
	
	\node[ll,input] at (labelpos-|input2) {Input};
	\node[ll,big] at (labelpos-|out1) {Ours-17-NoSM};
	\node[ll,full] at (labelpos-|out2) {Ground Truth};
	\node[ll,gt] at (labelpos-|out3) {3D Photo~\citep{3D_photo}};
	\node[ll,p3d] at (labelpos-|out4) {Stereo-Mag~\cite{stereo_mag}};
	\end{tikzpicture}
	\caption{Extrapolation comparison of our method ablation Ours-17-NoSM without soft masks against ground truth, 3D Photo~\citep{3D_photo}, and Stereo-Mag~\citep{stereo_mag}.
		The input frame closer to the target frame is marked in green for easier comparison.}
    \label{fig:full_generalized}
    
\end{figure*}

\tikzstyle{igt}=[fill=green!40,ultra thick]
\tikzstyle{ipd}=[fill=blue!50]
\tikzstyle{iarr}=[line width=4pt,draw=blue!30]

\begin{figure*} \centering
	\setlength{\imgheight}{2.4cm}

	\begin{tikzpicture}[every node/.style={inner sep=2pt}, font=\sffamily,
	spy using outlines={magnification=3, size=1.5cm, connect spies,
		every spy on node/.append style={thick},
		spy connection path={\draw[thick] (tikzspyonnode) -- (tikzspyinnode);}
	}]
	\matrix (m) [matrix of nodes] {
		|[igt]| \includegraphics[height=\imgheight]{images/interpolation/134_ours32/helper_frame.png} &
		|[igt]| \includegraphics[height=\imgheight]{images/interpolation/134_ours32/target_frame_8.png} &
		|[igt]| \includegraphics[height=\imgheight]{images/interpolation/134_ours32/target_frame_15.png} &
		|[igt]| \includegraphics[height=\imgheight]{images/interpolation/134_ours32/reference_frame.png} \\
		&
		|[ipd]| \includegraphics[height=\imgheight]{images/interpolation/134_ours32/8.png} &
		|[ipd]| \includegraphics[height=\imgheight]{images/interpolation/134_ours32/15.png} &
		& \\
	};
	\draw[iarr,-latex] (m-1-1) |- (m-2-2);
	\draw[iarr,latex-] (m-2-3) -| (m-1-4);
	\end{tikzpicture}
	
	\begin{tikzpicture}[every node/.style={inner sep=2pt}, font=\sffamily,
	spy using outlines={magnification=3, size=1.5cm, connect spies,
		every spy on node/.append style={thick},
		spy connection path={\draw[thick] (tikzspyonnode) -- (tikzspyinnode);}
	}]
	\matrix (m) [matrix of nodes] {
		|[igt]| \includegraphics[height=\imgheight]{images/interpolation/496_ours32/helper_frame.png} &
		|[igt]| \includegraphics[height=\imgheight]{images/interpolation/496_ours32/target_frame_5.png} &
		|[igt]| \includegraphics[height=\imgheight]{images/interpolation/496_ours32/target_frame_13.png} &
		|[igt]| \includegraphics[height=\imgheight]{images/interpolation/496_ours32/reference_frame.png} \\
		&
		|[ipd]| \includegraphics[height=\imgheight]{images/interpolation/496_ours32/5.png} &
		|[ipd]| \includegraphics[height=\imgheight]{images/interpolation/496_ours32/13.png} &
		& \\
	};
	\draw[iarr,-latex] (m-1-1) |- (m-2-2);
	\draw[iarr,latex-] (m-2-3) -| (m-1-4);
	\end{tikzpicture}
	
	\begin{tikzpicture}[every node/.style={inner sep=2pt}, font=\sffamily,
	spy using outlines={magnification=3, size=1.5cm, connect spies,
		every spy on node/.append style={thick},
		spy connection path={\draw[thick] (tikzspyonnode) -- (tikzspyinnode);}
	}]
	\matrix (m) [matrix of nodes] {
		|[igt]| \includegraphics[height=\imgheight]{images/interpolation/1245_ours32/helper_frame.png} &
		|[igt]| \includegraphics[height=\imgheight]{images/interpolation/1245_ours32/target_frame_4.png} &
		|[igt]| \includegraphics[height=\imgheight]{images/interpolation/1245_ours32/target_frame_9.png} &
		|[igt]| \includegraphics[height=\imgheight]{images/interpolation/1245_ours32/reference_frame.png} \\
		&
		|[ipd]| \includegraphics[height=\imgheight]{images/interpolation/1245_ours32/4.png} &
		|[ipd]| \includegraphics[height=\imgheight]{images/interpolation/1245_ours32/9.png} &
		& \\
	};
	\draw[iarr,-latex] (m-1-1) |- (m-2-2);
	\draw[iarr,latex-] (m-2-3) -| (m-1-4);
	\end{tikzpicture}
	
	\begin{tikzpicture}[every node/.style={inner sep=2pt}, font=\sffamily,
	spy using outlines={magnification=3, size=1.5cm, connect spies,
		every spy on node/.append style={thick},
		spy connection path={\draw[thick] (tikzspyonnode) -- (tikzspyinnode);}
	}]
	\matrix (m) [matrix of nodes] {
		|[igt]| \includegraphics[height=\imgheight]{images/interpolation/6841_ours32/helper_frame.png} &
		|[igt]| \includegraphics[height=\imgheight]{images/interpolation/6841_ours32/target_frame_5.png} &
		|[igt]| \includegraphics[height=\imgheight]{images/interpolation/6841_ours32/target_frame_13.png} &
		|[igt]| \includegraphics[height=\imgheight]{images/interpolation/6841_ours32/reference_frame.png} \\
		&
		|[ipd]| \includegraphics[height=\imgheight]{images/interpolation/6841_ours32/5.png} &
		|[ipd]| \includegraphics[height=\imgheight]{images/interpolation/6841_ours32/13.png} &
		& \\
	};
	\draw[iarr,-latex] (m-1-1) |- (m-2-2);
	\draw[iarr,latex-] (m-2-3) -| (m-1-4);
	\end{tikzpicture}
	\caption{Interpolation using the fast Ours-19 network. The network runs inference in 576p while being trained in 288p. In every block, the top row (green) shows the ground truth trajectory from RealEstate10k, while the bottom row (blue) presents the interpolated result corresponding to the ground truth camera poses.}
	\label{fig:interpolation}
\end{figure*}

\begin{figure*} \centering
 \setlength{\imgheight}{2.4cm}
 \begin{tikzpicture}[every node/.style={inner sep=2pt}, font=\sffamily,
    row 1 column 1/.style={fill=red,draw=red}]
  \matrix (m) [matrix of nodes] {
    |[igt]| \includegraphics[height=\imgheight]{images/interpolation/121_ours19/helper_frame.png} &
    |[igt]| \includegraphics[height=\imgheight]{images/interpolation/121_ours19/target_frame_7.png} &
    |[igt]| \includegraphics[height=\imgheight]{images/interpolation/121_ours19/target_frame_11.png} &
    |[igt]| \includegraphics[height=\imgheight]{images/interpolation/121_ours19/reference_frame.png} \\
    &
    |[ipd]| \includegraphics[height=\imgheight]{images/interpolation/121_ours19/7.png} &
    |[ipd]| \includegraphics[height=\imgheight]{images/interpolation/121_ours19/11.png} &
    & \\
  };
  \draw[iarr,-latex] (m-1-1) |- (m-2-2);
  \draw[iarr,latex-] (m-2-3) -| (m-1-4);
 \end{tikzpicture}

 \begin{tikzpicture}[every node/.style={inner sep=2pt}, font=\sffamily,
    spy using outlines={magnification=3, size=1.5cm, connect spies,
      every spy on node/.append style={thick},
      spy connection path={\draw[thick] (tikzspyonnode) -- (tikzspyinnode);}
    }]
  \matrix (m) [matrix of nodes] {
    |[igt]| \includegraphics[height=\imgheight]{images/interpolation/130_ours19/helper_frame.png} &
    |[igt]| \includegraphics[height=\imgheight]{images/interpolation/130_ours19/target_frame_7.png} &
    |[igt]| \includegraphics[height=\imgheight]{images/interpolation/130_ours19/target_frame_14.png} &
    |[igt]| \includegraphics[height=\imgheight]{images/interpolation/130_ours19/reference_frame.png} \\
    &
    |[ipd]| \includegraphics[height=\imgheight]{images/interpolation/130_ours19/7.png} &
    |[ipd]| \includegraphics[height=\imgheight]{images/interpolation/130_ours19/14.png} &
    & \\
  };
  \draw[iarr,-latex] (m-1-1) |- (m-2-2);
  \draw[iarr,latex-] (m-2-3) -| (m-1-4);
 \end{tikzpicture}

 \begin{tikzpicture}[every node/.style={inner sep=2pt}, font=\sffamily,
    spy using outlines={magnification=3, size=1.5cm, connect spies,
      every spy on node/.append style={thick},
      spy connection path={\draw[thick] (tikzspyonnode) -- (tikzspyinnode);}
    }]
  \matrix (m) [matrix of nodes] {
    |[igt]| \includegraphics[height=\imgheight]{images/interpolation/1350_ours19/helper_frame.png} &
    |[igt]| \includegraphics[height=\imgheight]{images/interpolation/1350_ours19/target_frame_7.png} &
    |[igt]| \includegraphics[height=\imgheight]{images/interpolation/1350_ours19/target_frame_22.png} &
    |[igt]| \includegraphics[height=\imgheight]{images/interpolation/1350_ours19/reference_frame.png} \\
    &
    |[ipd]| \includegraphics[height=\imgheight]{images/interpolation/1350_ours19/7.png} &
    |[ipd]| \includegraphics[height=\imgheight]{images/interpolation/1350_ours19/22.png} &
    & \\
  };
  \draw[iarr,-latex] (m-1-1) |- (m-2-2);
  \draw[iarr,latex-] (m-2-3) -| (m-1-4);
 \end{tikzpicture}

 \begin{tikzpicture}[every node/.style={inner sep=2pt}, font=\sffamily,
    spy using outlines={magnification=3, size=1.5cm, connect spies,
      every spy on node/.append style={thick},
      spy connection path={\draw[thick] (tikzspyonnode) -- (tikzspyinnode);}
    }]
  \matrix (m) [matrix of nodes] {
    |[igt]| \includegraphics[height=\imgheight]{images/interpolation/3527_ours19/helper_frame.png} &
    |[igt]| \includegraphics[height=\imgheight]{images/interpolation/3527_ours19/target_frame_5.png} &
    |[igt]| \includegraphics[height=\imgheight]{images/interpolation/3527_ours19/target_frame_15.png} &
    |[igt]| \includegraphics[height=\imgheight]{images/interpolation/3527_ours19/reference_frame.png} \\
    &
    |[ipd]| \includegraphics[height=\imgheight]{images/interpolation/3527_ours19/5.png} &
    |[ipd]| \includegraphics[height=\imgheight]{images/interpolation/3527_ours19/15.png} &
    & \\
  };
  \draw[iarr,-latex] (m-1-1) |- (m-2-2);
  \draw[iarr,latex-] (m-2-3) -| (m-1-4);
 \end{tikzpicture}
 \caption{Interpolation using the Ours-32 network. The network runs inference in 576p while being trained in 288p. In every block, the top row (green) shows the ground truth trajectory from RealEstate10k, while the bottom row (blue) presents the interpolated result corresponding to the ground truth camera poses.}
 \label{fig:interpolation2}
\end{figure*}

 \fi

\end{document}